\documentclass{article}

 \usepackage[preprint]{neurips_2026}

\usepackage[utf8]{inputenc} 
\usepackage[T1]{fontenc}    
\usepackage[hidelinks,colorlinks]{hyperref}       
\usepackage{url}            
\usepackage{booktabs}       
\usepackage{amsfonts}       
\usepackage{nicefrac}       
\usepackage{microtype}      
\usepackage{xcolor}         
\usepackage{graphicx}

\usepackage{amsmath}
\usepackage{multirow}
\usepackage[capitalize,noabbrev]{cleveref}

\usepackage{pifont}
\usepackage{placeins}

\usepackage{enumitem}
\setlist{leftmargin=2em,topsep=0pt,partopsep=0pt,parsep=0pt,itemsep=0pt}

\definecolor{myblue}{rgb}{0,0.3,0.6}
\hypersetup{linkcolor=myblue,urlcolor=myblue,citecolor=myblue,anchorcolor=myblue}

\newcommand{\yes}{\textcolor{green!50!black}{\ding{51}}}
\newcommand{\no}{\textcolor{red!70!black}{\ding{55}}}

\crefformat{footnote}{#2\footnotemark[#1]#3}
\crefrangeformat{footnote}{#3\footnotemark[#1]#4--#5\footnotemark[#2]#6}
\crefmultiformat{footnote}{#2\footnotemark[#1]#3}{\textsuperscript{,}#2\footnotemark[#1]#3}{\textsuperscript{,}#2\footnotemark[#1]#3}{\textsuperscript{,}#2\footnotemark[#1]#3}

\title{Clips2Story: Training-Free Video Storyboarding and Editing Using Multimodal Retrieval-Embedded Generation}

\author{%
  Junnuo Wang \\
  New York University \\
  \href{mailto:jw6641@nyu.edu}{\texttt{jw6641@nyu.edu}}\\
  \And
  Weihan Xu \\
  University of Washington \\
  \href{mailto:wx35@uw.edu}{\texttt{wx35@uw.edu}}\\
  \And
  Hao-Wen Dong \\
  University of Michigan \\
  \href{mailto:hwdong@umich.edu}{\texttt{hwdong@umich.edu}}
}

\begin{document}

\maketitle

\begin{abstract}
Video editors often work with large collections of fragmented footage and must select and organize relevant clips into a coherent story. However, many existing video editing and summarization methods assume a temporally ordered source video and aim to preserve its original narrative and temporal structure. In this paper, we introduce \textit{Clips2Story}, a training-free, LLM-based pipeline that enables users to create keyword-specified short videos from a pool of video clips. Given a pool of clips and a user-specified keyword, we first extract text-based multimodal descriptors for each clip and retrieve the clips most relevant to the keyword. We then provide the textual descriptors of these candidate clips to an LLM, which selects and arranges them into an editable storyboard with connective narrations. Finally, we assemble the selected clips and retrieve accompanying visuals for the narrations to compose the output video. To evaluate the proposed pipeline, we construct a synthetic evaluation dataset from 100 long-form videos across five genres by segmenting each video into short clips using a scene-change detection model and shuffling the resulting clips to remove their temporal order. Through objective evaluation and qualitative examples, we demonstrate that Clips2Story can effectively retrieve keyword-relevant clips and compose coherent storyboards based on user-specified keywords. Human evaluation further shows that Clips2Story outperforms a strong baseline in terms of narrative coherence, keyword alignment, and overall realism.
\end{abstract}

\section{Introduction}
\label{sec:intro}

Video editing is a laborious yet critical task for video production. Video editors oftentimes need to look at a large amount of raw video footage, select useful video clips, and arrange them into the desired narrative. Yet, existing video summarization models require a long input video with a coherent storyline \citep{lin2023videoxum}; existing long-form video generation models cannot retrieve and integrate raw video footage into their outputs; existing video editing approaches can only generate videos that follow the exact storyline of the input video \citep{xu2025regen}.
They are therefore not designed for a common editing problem: constructing a new, user-directed story from a repository of existing video materials.

In this work, we aim to explore AI-assisted video editing that can 1)
work with real video footage, 2) adapt to the user's creative intention,
and 3) generate a human-editable intermediate output that allows human intervention.
Among the many complex tasks involved in professional video editing workflows,
we focus on two central steps: selecting relevant video clips and organizing the selected video clips into a narrative specified by a keyword provided by the user.
We introduce the \textit{clips-to-storyboard} generation task and present \textit{Clips2Story}, a training-free, LLM-based framework that takes a pool of video clips and a user-specified keyword as input. We first extract text-based descriptors for each video clip using speech recognition, video captioning, and object detection models. Next, we retrieve the video clips that are most relevant to the keyword provided by the user. Finally, we instruct a large language model (LLM) to arrange them into a coherent storyboard based on the multimodal information extracted from the candidate video clips.
Unlike conventional video summarization models that produce a fixed summary that follows the original storyline, Clips2Story can generate diverse storyboards that differ from the input video based on user-specified keywords.

\begin{figure*}[t]
\centering
\includegraphics[width=\textwidth]{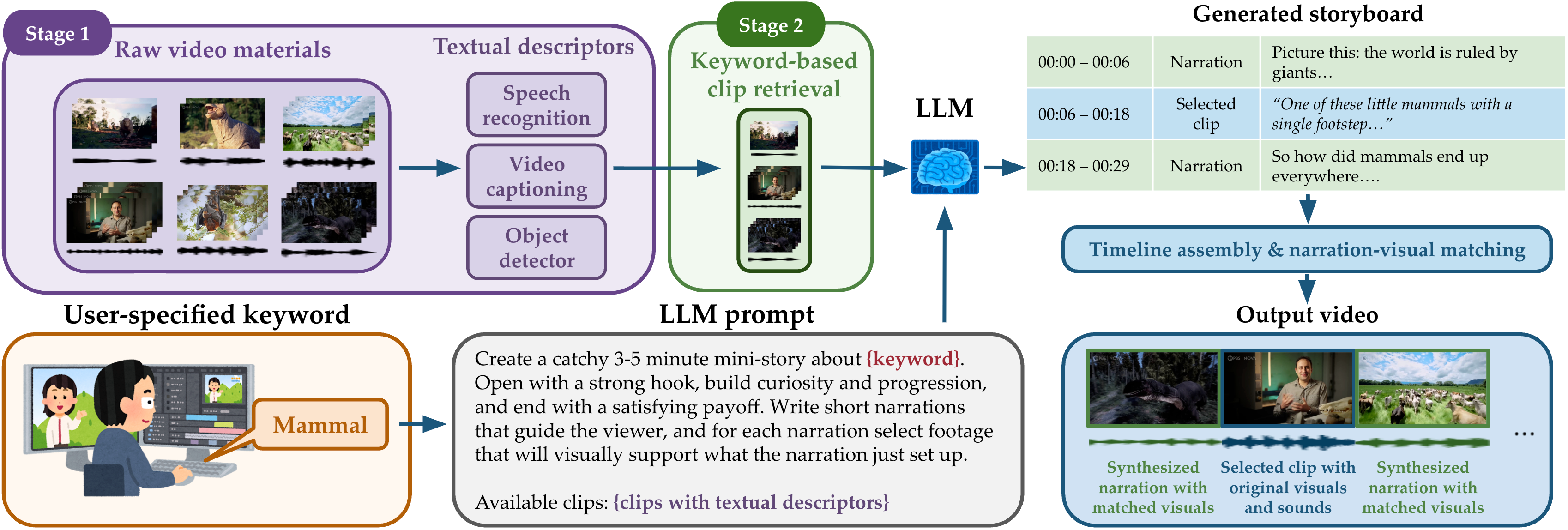}
\vspace{-3ex}
\caption{Overview of the proposed Clips2Story-ND pipeline. Given a pool of raw video footage and a user-specified keyword, we first extract textual descriptors from the video clips and retrieve those most relevant to the keyword. We then use an LLM to generate an editable storyboard with connective narration. Finally, we compose the output video by assembling the selected clips and retrieving accompanying visuals for the narration.
}
\label{fig:clips2story_nd}
\vspace{-1ex}
\end{figure*}

With the generated storyboard, we then assemble the selected clips to compose the final output video. We study two variants: the \textit{Clips2Story-NF} model sequences existing clips without connective narration; the \textit{Clips2Story-ND} model generates connective narration to improve coherence between video clips. Rather than directly producing a finalized video, both Clips2Story variants output a human-readable storyboard containing the proposed clip sequence and narration, allowing creators to inspect and revise the narrative plan before final assembly.

To evaluate the proposed models, we construct a synthetic evaluation dataset from 100 long-form videos spanning five genres: documentary, film, lecture, vlog, and news. We first segment each video into short clips using a scene-change detection model at both the clip and scene levels. We then shuffle the resulting clips before sending their textual descriptors to the LLM, thereby removing the original temporal order and preventing leakage of the input video's storyline. Our goal is to approximate the fragmented organization of real footage repositories that video editors often work with. We evaluate Clips2Story using automatic metrics, human evaluation, and ablations. The metrics assess repetition, visual and keyword alignment, temporal granularity, and consistency with the source order as a diagnostic reference. Across both settings, Clips2Story-NF achieves the strongest order preservation and lower repetition than the baselines, while Clips2Story-ND achieves the strongest overall keyword alignment. For human evaluation, Clips2Story-ND significantly outperforms the baseline model in terms of coherence, keyword alignment, realness, narration effectiveness, and clip-selection effectiveness.

Our main contributions can be summarized as follows:
\begin{itemize}
\item We formulate keyword-guided \textit{clips-to-storyboard} generation as a task that aims to edit a set of video clips into a storyboard with a coherent narrative specified by the user, representing a step toward practical video editing with existing footage.

\item We introduce Clips2Story, a training-free framework that combines keyword-guided retrieval, multimodal descriptor extraction, and LLM-based storyboard planning. Its intermediate storyboard makes the proposed clip sequence and narration human-readable and editable.

\item We construct a synthetic evaluation dataset from 100 real long-form videos across five genres and develop clip-level and scene-level settings for evaluating storyboard generation from fragmented video materials.
\end{itemize}

All source code and implementation details will be released upon paper acceptance. Video samples can be found on our demo website.\footnote{\url{https://junnuo3.github.io/clips2story/}\label{fn:demo}}

\section{Related Work}
\label{sec:related_work}

\paragraph{LLM-based video editing.}

Recent studies have explored LLMs as active agents for video editing and storytelling. \citet{wang2024lave} introduced LAVE, an LLM-powered assistant for language-guided linear video editing. For automated narrative construction, \citet{xu2024teasergen} proposed TeaserGen, which generates narration and retrieves visually aligned clips to create documentary teasers. Extending this narration-driven paradigm, \citet{xu2025regen} introduced REGen, which inserts retrieved video clips into explicit placeholders in an LLM-generated script. Although these systems demonstrate the potential of narration-guided and retrieval-augmented video editing, they assume that the input footage is chronologically ordered. In contrast, our method operates on a fragmented and randomly ordered pool of candidate clips, requiring the model to construct a globally coherent editorial structure from temporally unordered multimodal evidence, as summarized in \cref{app:comparisons}.

\paragraph{Video summarization.}

As a special case of video editing, video summarization shares the objective of distilling visual footage into a concise narrative with clips-to-storyboard generation. Existing approaches have used scene recognition and cross-modal alignment to generate condensed videos and textual summaries \citep{bose2023movieclip,lin2023videoxum,he2023align}. More recently, \citet{argaw2024scaling} leverage LLMs to extract salient sentences from video transcripts, while \citet{singh2024previously} use a hierarchical framework to identify sub-stories within structured television episodes. Unlike conventional video summarization, which condenses an existing narrative, our work composes a new storyline from a pool of candidate video clips.


\paragraph{LLM-based video generation.}

Recent advances have explored the use of LLMs for high-level semantic planning in video synthesis. \citet{lian2023llm} use LLMs to expand user prompts into structured multi-scene plans. Similarly, \citet{yeh2026vidmento} and \citet{kondratyuk2023videopoet} frame video authoring as a context-aware expansion process that bridges narrative gaps. These methods primarily focus on planning layouts or generating new visual content to realize predefined storylines. In contrast, Clips2Story formulates video creation as a multimodal editing and sequencing problem. Rather than generating new pixels, our method extracts rich textual descriptors from existing video assets and uses them as surrogate representations. This formulation allows the LLM to act as an editor by retrieving, arranging, and narrating existing content, thereby shifting the focus from pixel-level synthesis to high-level narrative assembly.

\section{Method}

As illustrated in \cref{fig:clips2story_nd}, our training-free, LLM-based video editing framework consists of two stages: (1) shot-level multimodal analysis and (2) keyword-based clip retrieval. These stages are followed by two alternative pipelines for planning and rendering: Clips2Story-NF (narration-free) and Clips2Story-ND (narration-driven).

\subsection{Shot-level Multimodal Analysis}

For each clip in the candidate pool, we first extract the transcript and speaker information using a pretrained automatic speech recognition model; our implementation adopts WhisperX \citep{bain2023whisperx}. We then select the middle frame of the clip and generate a visual caption using a vision-language foundation model, instantiated as BLIP \citep{li2022blip}.
Further, we extract entity and background information using an open-vocabulary object detector and a contrastive vision-language model, implemented with GroundingDINO \citep{liu2024grounding} and CLIP ViT-B/32 \citep{radford2021learning}, respectively. As a result, each clip is represented by a multimodal description containing its duration, transcript, caption, entities, and background cues. These textual descriptors serve as inputs for both semantic retrieval and downstream planning.


\subsection{Keyword-based Clip Retrieval}

A major challenge in AI-assisted video editing \citep{lin2023videodirectorgpt, wang2024lave} is scaling to a large number of input videos.
Directly feeding the multimodal descriptions of an entire film or documentary into an LLM overloads its effective reasoning capabilities. Without filtering, the model struggles to manage the massive context \citep{liu2024lost},
resulting in poorly
sequenced timelines and generated videos.
To make the editing task tractable, we construct a planning pool by retrieving clips based on their semantic relevance to a user-specified keyword. Specifically, we rank clip relevance by computing the cosine similarity between the sentence-transformer embeddings \citep{reimers2019sentence} of the keyword and each clip's concatenated transcript and caption.
We then retrieve a subset of candidate clips with the highest similarity scores, up to a total duration of 15 minutes.
This subset forms the planning pool presented to the LLM for timeline planning, which is necessary for handling long-form content.

\subsection{Narration-Free Timeline Sequencing}

Using the keyword-specific clip pool, we prompt an LLM, instantiated as GPT-5.2~\citep{openai2025gpt52}, to generate a video timeline centered on the user-specified keyword. We first introduce the \textit{Clips2Story-NF} pipeline for timeline planning. In this pipeline, the LLM is tasked exclusively with selecting and sequencing footage segments from the retrieved clip pool. Clips2Story-NF produces timelines composed entirely of the original audiovisual footage, without generating additional media such as synthetic narration. It is therefore suitable for lightweight assistance with organizing raw footage or drafting storyboards. The prompt used for Clips2Story-NF is provided in Appendix~\ref{app:nf_prompt}.

\subsection{Narration-Driven Visual Selection and Rendering}

Although Clips2Story-NF sequences relevant clips, directly assembling disjointed segments can result in abrupt transitions and weak logical flow. To bridge semantic gaps between clips, we introduce \textit{Clips2Story-ND}. As illustrated in \cref{fig:clips2story_nd}, we instruct the LLM to generate a sequence of short connective narrations that support the selected video clips. We then apply a visual selection stage to determine which footage should be displayed during each generated narration segment. Following \citet{xu2024teasergen}, we synthesize the generated narration segments into audio using a text-to-speech model; our implementation uses VITS~\citep{kim2021conditional}. We then extract frames from the candidate clip pool and use a VTGHLS-based score to rank and select visually relevant footage whose duration matches that of the corresponding narration.

The final Clips2Story-ND video combines three components: (1) synthesized narration audio; (2) retrieved clips played under the narration with their original audio muted; and (3) LLM-selected clips that retain both their original audio and visual content.
Clips2Story-ND is suitable for editing scenarios in which the source footage alone is insufficient to produce a longer video with a coherent narrative, such as explanatory videos or short stories assembled from heterogeneous raw clips.
We provide the specific prompt used for Clips2Story-ND in \cref{app:nd_prompt}. 

\begin{table*}[t]
    \centering
    \small
    \caption{Statistics of the evaluation datasets.}
    \label{tab:duration_stats}
    \vspace{1ex}
    \begin{tabular}{
        l@{~~~~~}ccc@{~~}cc@{~~}c
    }
        \toprule
        \multirow{2}{*}[-2pt]{Genre}
        & \multirow{2}{*}[-3pt]{%
            \shortstack{Video duration (min)\\
            (mean $\pm$ std. dev.)}}
        & \multirow{2}{*}[-3pt]{%
            \shortstack{Word density\\
            (\# words/min)}}
        & \multicolumn{2}{c}{Clip-level shuffling}
        & \multicolumn{2}{c}{Scene-level shuffling}\\[-1pt]
        \cmidrule(lr){4-5}
        \cmidrule(lr){6-7}
        &
        &
        & Count
        & Duration (sec)
        & Count
        & Duration (sec) \\
        \midrule
        Film
        & $86.45 \pm 13.31$
        & 152.43
        & 656.55
        & 9.66
        & 164.95
        & 38.33 \\

        Documentary
        & $48.43 \pm 4.68$
        & 161.27
        & 730.75
        & 4.13
        & 183.55
        & 16.42 \\

        Lecture
        & $41.81 \pm 11.25$
        & 168.50
        & 99.05
        & 165.22
        & 25.70
        & 300.32 \\

        Vlog
        & $27.04 \pm 6.35$
        & 186.46
        & 52.65
        & 74.36
        & 14.05
        & 198.39 \\

        News
        & $24.05 \pm 4.74$
        & 187.95
        & 198.65
        & 8.44
        & 50.50
        & 32.99 \\
        \midrule
        All
        & $45.56 \pm 8.07$
        & 171.32
        & 347.53
        & 52.36
        & 87.75
        & 117.29 \\
        \bottomrule
    \end{tabular}
\end{table*}

\section{Experimental Setup}
\label{sec:setup}

\subsection{Evaluation Dataset}
\label{sec:dataset}

Because paired data for clips-to-storyboard generation are unavailable, we construct a controlled evaluation dataset from 100 long-form YouTube videos, evenly distributed across five genres: documentary, film, lecture, vlog, and news. We segment each source video into candidate units and shuffle them before inference to approximate real-world editing scenarios, in which footage is often stored as fragmented, unordered material rather than presented as a coherent sequence. This removes direct access to the original chronology while retaining it as a diagnostic reference. As summarized in \cref{tab:duration_stats}, the genres vary substantially in video duration and word density, allowing evaluation across different temporal structures and content densities.

\begin{figure*}[t]
    \centering
    \includegraphics[width=\textwidth]{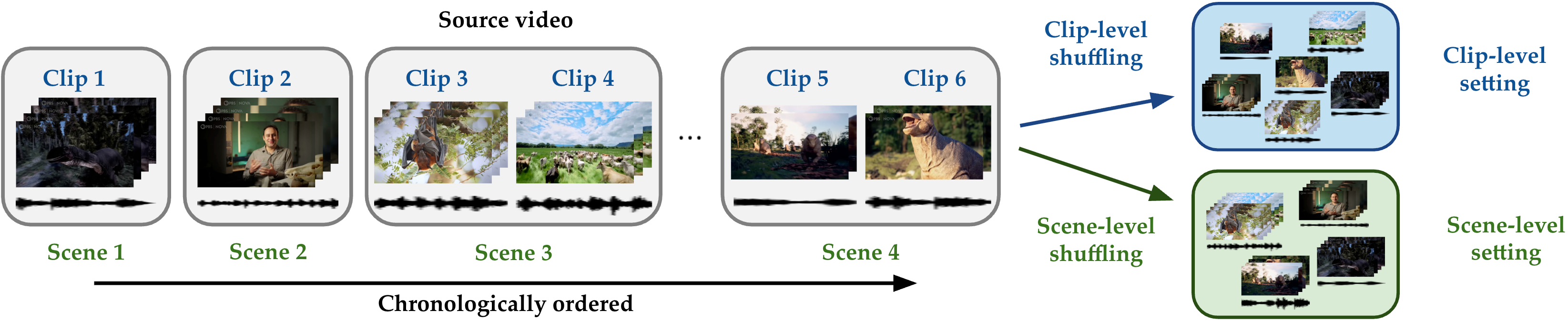}
    \caption{Illustration of the dataset construction process. The original input video is processed into two distinct candidate pools: a fine-grained clip-level pool and a grouped scene-level pool.}
    \label{fig:dataset_buildup}
\end{figure*}

We construct two versions of the dataset, as illustrated in \cref{fig:dataset_buildup}, to evaluate performance under different levels of footage fragmentation.

\begin{itemize}
\item \noindent\textit{Clip-level setting}:
We divide each source video into short shots using the automated transition detector TransNetV2~\citep{soucek2024transnet} and shuffle the resulting clips. The large pool of short, fragmented candidates provides limited temporal context and constitutes the more challenging setting.

\item \noindent\textit{Scene-level setting}:
To approximate footage in which multiple camera shots capture the same semantic event~\citep{bose2023movieclip}, we merge adjacent shots using a multimodal similarity score based on visual, textual, and audio features, as described in \cref{app:dataset_details}. This produces fewer, longer candidate units with greater temporal context and therefore represents a less fragmented setting. Together, these settings provide a controlled means of examining how candidate granularity affects keyword-guided storyboard planning.

\end{itemize}

\subsection{Keyword Prompt Generation}

To simulate user-directed editing, we use an LLM, instantiated as GPT-5.2~\citep{openai2025gpt52}, to generate five representative keywords for each source video based on its full transcript and captions. Each keyword is incorporated into a shared prompt template, allowing us to evaluate whether the systems can produce distinct, semantically focused storyboards from the same candidate footage.

\subsection{Baseline Models}

We compare our method against three baseline models. For consistency, all baselines are adapted to operate on the same shuffled candidate pool as our method.

\begin{itemize}

\item \noindent\textit{A2Summ} \citep{he2023align} is an extractive multimodal video summarization model that selects segments based on text--vision temporal correspondence. To accommodate long inputs, we apply it to 300-second chunks and concatenate the resulting selections.

\item \noindent\textit{TeaserGen} \citep{xu2024teasergen} is a narration-centric, LLM-based framework for teaser generation. It divides the transcript into ten segments, summarizes and rewrites them into a cohesive narration, and retrieves visual clips aligned with the synthesized audio.

\item \noindent\textit{REGen} \citep{xu2025regen} is a retrieval-embedded framework for long-to-short video editing. It generates a narration script containing visual placeholders and fills them with retrieved clips to produce the final video.

\end{itemize}

\subsection{Evaluation Metrics}
\label{sec}
Following TeaserGen \citep{xu2024teasergen} and REGen \citep{xu2025regen}, we report Repetitiveness (REP), CLIPScore, and VTGHLS. To further measure
the semantic alignment and temporal structure of the generated storyboards, we introduce three task-specific metrics:

\begin{itemize}
    
\item \noindent\textit{Mean clip duration} (\textbf{MCD})
is the average duration of the selected clips and generated narration segments, capturing
the pacing and temporal granularity of the final sequence.

\item \noindent\textit{Order preservation coefficient} (\textbf{OPC})
measures the consistency between the generated clip order and the original chronological order of the source video using Kendall's $\tau$. Although our task does not aim to reconstruct the source chronology, the original order provides an informative proxy for natural temporal and narrative progression. A higher OPC therefore indicates that the generated timeline more closely follows the underlying narrative flow of the source video.

\item \noindent\textit{Keyword relevance} (\textbf{KR})
measures the semantic alignment between the generated timeline and its target keyword. We compute three modality-specific scores based on (1) narration and transcripts, (2) video frames, and (3) visual captions. Each modality is divided into approximately uniform textual or temporal units, and its KR score is computed as the mean cosine similarity between the keyword and the corresponding units.

\end{itemize}
Formal definitions and implementation details are provided in \cref{app:eval_metrics}.

\subsection{Subjective Evaluation}

We conduct a human evaluation with 15 participants using a 5-point Likert scale. We consider three genres: documentary, film, and lecture. For each genre, we randomly sample six videos and ask participants to evaluate the short videos (3--5 minutes) generated by Clips2Story-NF, Clips2Story-ND, and REGen~\citep{xu2025regen}. The study evaluates the generated videos across five dimensions: coherence, keyword alignment, realness, narration effectiveness, and clip-selection effectiveness. Additionally, we compare the textual storyboards generated by REGen and Clips2Story-ND across three dimensions: organization, keyword informativeness, and engagingness. The exact questionnaire items are provided in \cref{app:subjective_video_questions,app:subjective_narration_questions}.


\section{Results}
\label{sec:results}

\subsection{Qualitative Examples}

\begin{figure*}[t]
    \centering
    \begin{minipage}[t]{0.42\textwidth}
        \centering
        \includegraphics[width=\linewidth]{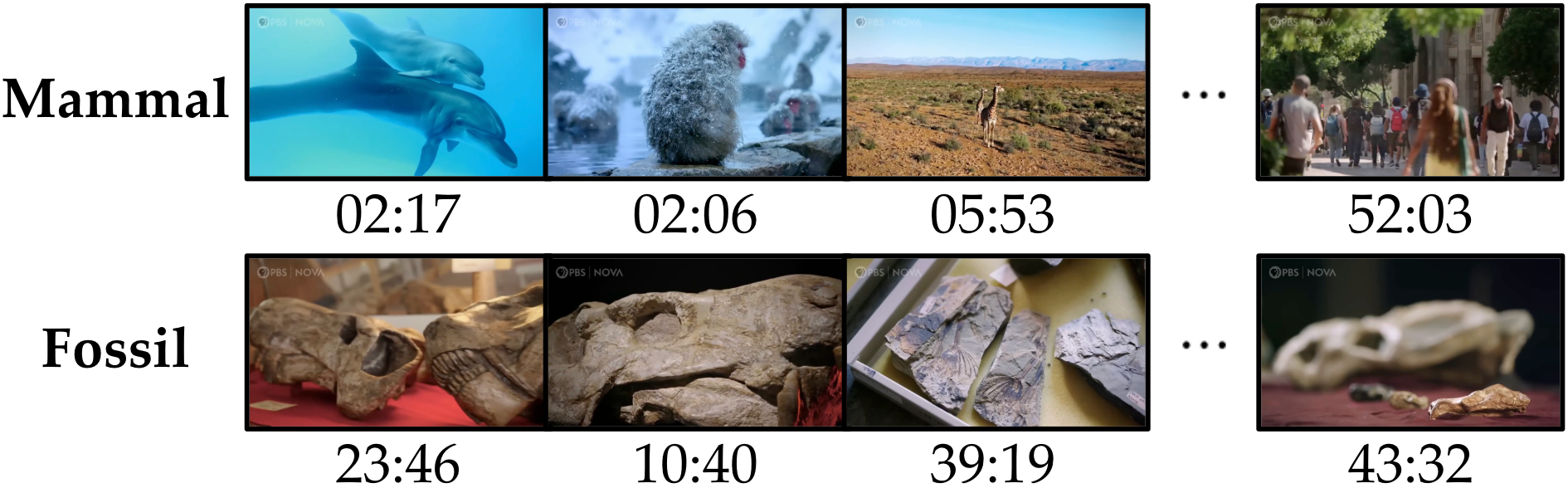}
    \end{minipage}
    \begin{minipage}[t]{0.28\textwidth}
        \centering
        \includegraphics[width=\linewidth]{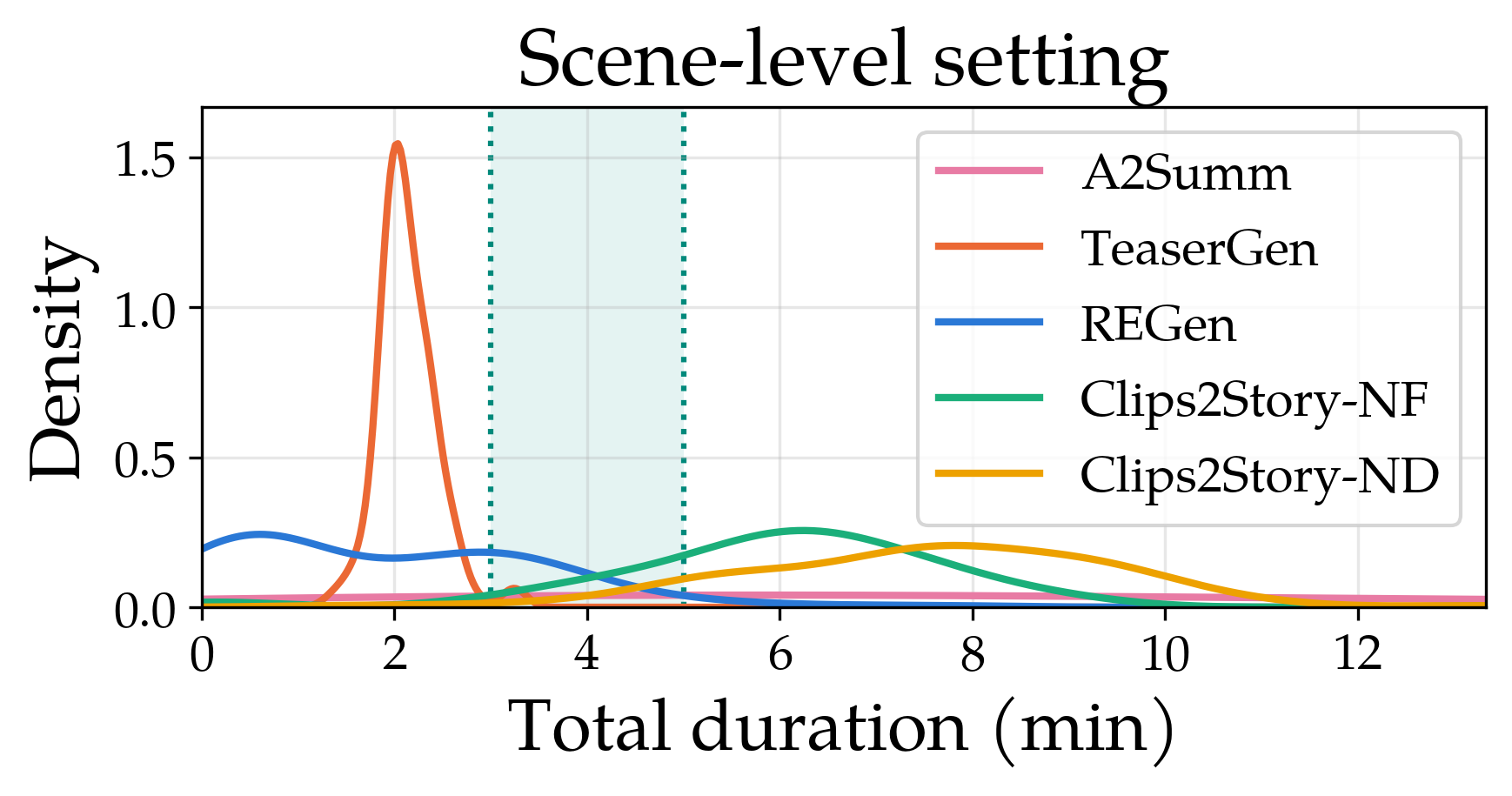}
    \end{minipage}
    \begin{minipage}[t]{0.28\textwidth}
        \centering
        \includegraphics[width=\linewidth]{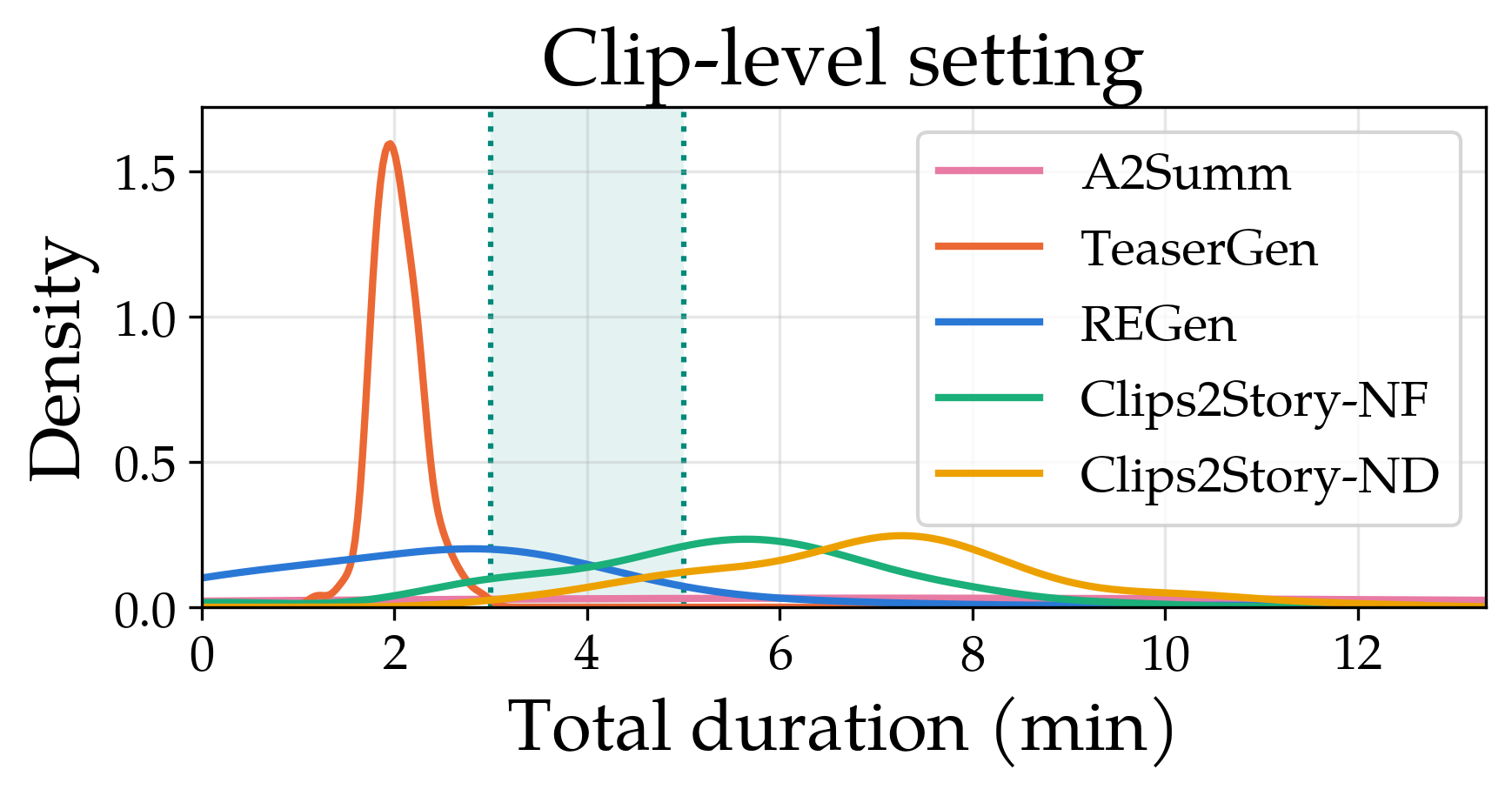}
    \end{minipage}
    \caption{Qualitative storyboards generated by Clips2Story-ND, with each clip labeled by its original start time in the source video (left), alongside generated storyboard duration distributions under the scene-level (middle) and clip-level (right) settings.
}
    \label{fig:qualitative}
\end{figure*}

\cref{fig:qualitative} presents two example storyboards generated by Clips2Story-ND for the same documentary under the scene-level setting, conditioned on the keywords ``mammal'' and ``fossil.'' The selected clips exhibit clear semantic alignment with their respective keywords. For ``mammal,'' the storyboard includes dolphins, primates, and humans, whereas ``fossil'' retrieves footage of fossilized skulls and bones. Furthermore, the timestamps below the clips denote their original start times in the source video. We observe that the generated storyboards can reconstruct parts of the original timeline while preserving narrative and logical flow. These examples demonstrate the system's ability to customize storyboards according to user-specified keywords while organizing semantically related clips into coherent sequences.


\begin{table*}[t]
    \centering
    \footnotesize
    \setlength{\tabcolsep}{3pt}
    \caption{
        Objective evaluation results under the scene-level and clip-level
        settings. Error bars indicate 95\% confidence intervals.
    }
    \label{tab:objective_combined}
    \vspace{1ex}
    \resizebox{\textwidth}{!}{%
        \begin{tabular}{llcccccccc}
            \toprule
            &Model
            & MCD (sec)
            & REP (\%)
            & CLIPScore
            & VTGHLS
            & OPC
            & KR-Narration $\uparrow$
            & KR-Frame $\uparrow$
            & KR-Caption $\uparrow$ \\
            \midrule
            

            \multirow{5}{*}{\rotatebox[origin=c]{90}{\shortstack{Scene-level\\shuffling}}}
            &A2Summ {\scriptsize\citep{he2023align}}
            & $15.58 \pm 3.86$
            & $0.00 \pm 0.00$
            & $0.54 \pm 0.01$
            & $1.04 \pm 0.02$
            & $0.18 \pm 0.06$
            & $0.16 \pm 0.01$
            & $0.17 \pm 0.00$
            & $0.21 \pm 0.01$\\

            &TeaserGen {\scriptsize\citep{xu2024teasergen}}
            & $4.19 \pm 0.15$
            & $18.95 \pm 2.29$
            & $0.60 \pm 0.02$
            & $1.41 \pm 0.02$
            & $0.05 \pm 0.05$
            & $0.18 \pm 0.01$
            & $0.18 \pm 0.01$
            & $0.22 \pm 0.02$ \\

            &REGen {\scriptsize\citep{xu2025regen}}
            & $4.85 \pm 0.90$
            & $20.39 \pm 6.07$
            & $0.58 \pm 0.02$
            & $0.68 \pm 0.10$
            & $0.11 \pm 0.11$
            & $0.18 \pm 0.02$
            & $0.18 \pm 0.01$
            & $0.23 \pm 0.02$ \\

            \cmidrule(lr){2-10}

            &Clips2Story-NF
            & $28.75 \pm 2.67$
            & $0.89 \pm 0.46$
            & $0.56 \pm 0.02$
            & $1.06 \pm 0.03$
            & $0.56 \pm 0.04$
            & $0.22 \pm 0.01$
            & $\mathbf{0.19 \pm 0.00}$
            & $0.23 \pm 0.02$ \\

            &Clips2Story-ND
            & $10.59 \pm 0.79$
            & $13.11 \pm 2.14$
            & $0.56 \pm 0.01$
            & $1.20 \pm 0.01$
            & $0.27 \pm 0.04$
            & $\mathbf{0.25 \pm 0.01}$
            & $\mathbf{0.19 \pm 0.00}$
            & $\mathbf{0.24 \pm 0.01}$ \\

            \cmidrule(lr){1-10}

            \multirow{5}{*}{\rotatebox[origin=c]{90}{\shortstack{Clip-level\\shuffling}}}
            &A2Summ {\scriptsize\citep{he2023align}}
            & $16.76 \pm 3.96$
            & $0.00 \pm 0.00$
            & $0.55 \pm 0.01$
            & $1.05 \pm 0.02$
            & $0.05 \pm 0.04$
            & $0.17 \pm 0.01$
            & $0.17 \pm 0.00$
            & $0.21 \pm 0.01$ \\

            &TeaserGen {\scriptsize\citep{xu2024teasergen}}
            & $4.11 \pm 0.14$
            & $16.86 \pm 2.44$
            & $0.60 \pm 0.01$
            & $1.41 \pm 0.02$
            & $0.07 \pm 0.06$
            & $0.18 \pm 0.01$
            & $0.18 \pm 0.01$
            & $0.23 \pm 0.02$ \\

            &REGen {\scriptsize\citep{xu2025regen}}
            & $4.90 \pm 0.50$
            & $20.21 \pm 6.38$
            & $0.58 \pm 0.02$
            & $0.78 \pm 0.10$
            & $0.07 \pm 0.08$
            & $0.18 \pm 0.02$
            & $\mathbf{0.19 \pm 0.01}$
            & $0.23 \pm 0.02$ \\

            \cmidrule(lr){2-10}

            &Clips2Story-NF
            & $15.22 \pm 2.19$
            & $0.66 \pm 0.29$
            & $0.55 \pm 0.02$
            & $1.05 \pm 0.03$
            & $0.30 \pm 0.06$
            & $0.24 \pm 0.02$
            & $\mathbf{0.19 \pm 0.00}$
            & $0.23 \pm 0.02$ \\

            &Clips2Story-ND
            & $7.64 \pm 0.74$
            & $15.80 \pm 2.05$
            & $0.55 \pm 0.01$
            & $1.24 \pm 0.01$
            & $0.15 \pm 0.04$
            & $\mathbf{0.26 \pm 0.01}$
            & $\mathbf{0.19 \pm 0.00}$
            & $\mathbf{0.24 \pm 0.01}$ \\
            \cmidrule(lr){1-10}

            &Input video clips
            & --
            & $0.02 \pm 0.00$
            & $0.50 \pm 0.01$
            & $1.00 \pm 0.02$
            & --
            & $0.15 \pm 0.01$
            & $0.17 \pm 0.00$
            & $0.20 \pm 0.01$ \\

            \bottomrule
        \end{tabular}%
    }
\end{table*}

\subsection{Objective Evaluation Results}

\cref{tab:objective_combined} presents the overall performance of our proposed pipelines and the baselines under both the scene-level and clip-level settings.

\paragraph{Narrative coherence and redundancy.}

Although explicit temporal reconstruction is not the primary objective of our task, the chronological order of the source video serves as a proxy for natural narrative progression. As shown in \cref{tab:objective_combined}, Clips2Story-NF achieves the highest OPC in both settings, suggesting that LLM-based planning can recover a coherent narrative trajectory without access to explicit temporal priors. In comparison, the three baselines, A2Summ, TeaserGen, and REGen, obtain OPC scores below 0.20. Moreover, TeaserGen and REGen exhibit higher repetitiveness (REP), indicating that they reuse the same content more frequently. Both Clips2Story variants achieve lower REP values than these baselines. Between our two variants, the narration-free Clips2Story-NF produces substantially less repetition than the narration-driven Clips2Story-ND. One possible explanation is that the narration-to-visual matching stage in Clips2Story-ND, which follows the retrieval approach introduced by TeaserGen, may retrieve the same visually relevant clips for semantically similar narration segments.

\begin{figure}[t]
    \centering
    \includegraphics[width=0.495\linewidth]
    {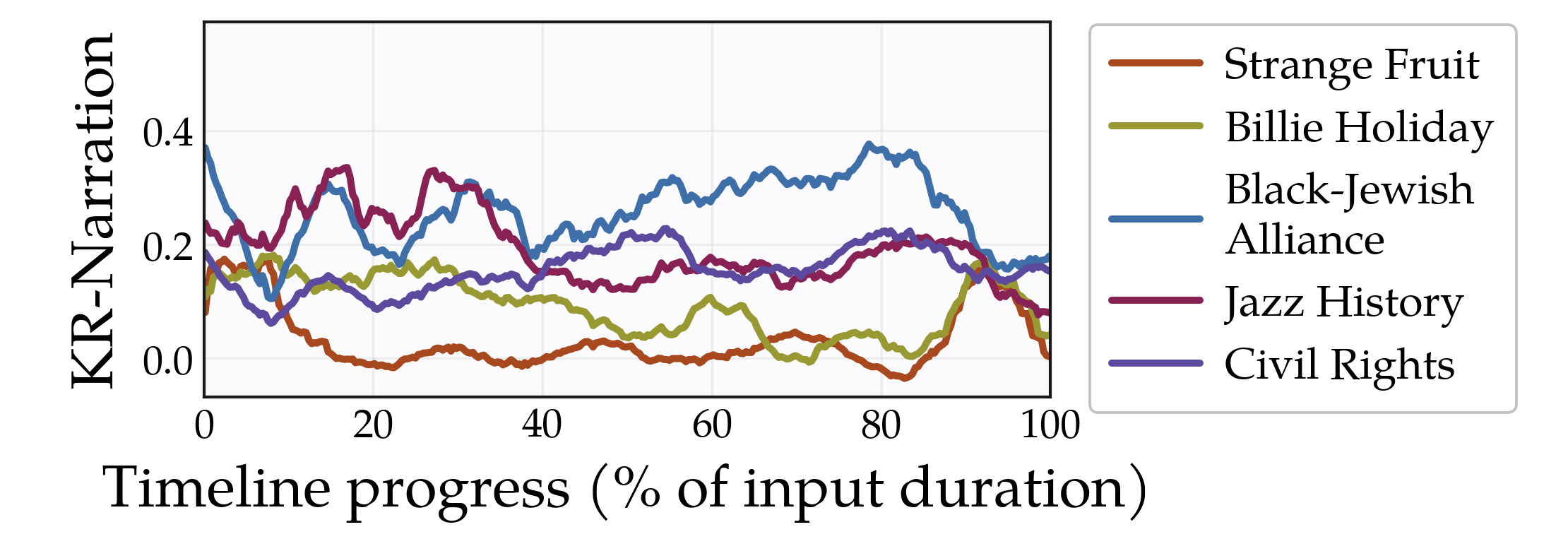}%
    \hfill
    \includegraphics[width=0.495\linewidth]
    {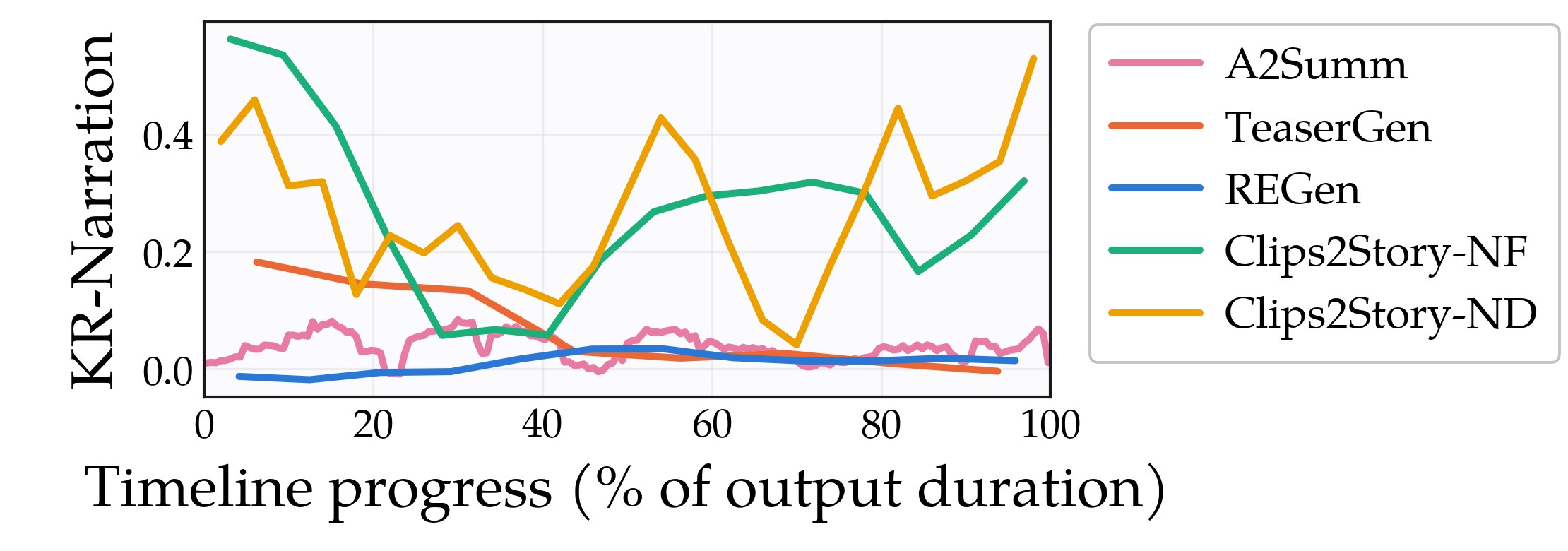}

    \caption{
    KR-Narration over the normalized timeline for the documentary
    \textit{Black and Jewish America: An Interwoven History}.
    Left: The input video evaluated with five target keywords.
    Right: Outputs generated by A2Summ, TeaserGen, REGen,
    Clips2Story-NF, and Clips2Story-ND for the target keyword
    ``Strange Fruit.''
    }
    \label{fig:keyword_relevance_timeline}
\end{figure}

\begin{figure}[t]
    \centering
    \includegraphics[width=\columnwidth]
    {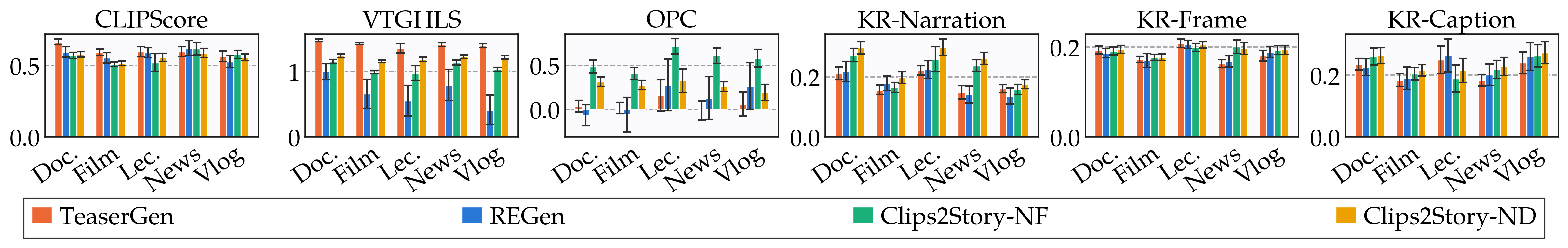}

    \caption{
    Genre-wise comparison of CLIPScore, VTGHLS, OPC, KR-Narration, KR-Frame, and KR-Caption under the scene-level setting. Error bars indicate 95\% confidence intervals. Specific numbers can be found in \cref{tab:cross_genre_scene_level}.
    }
    \label{fig:genre_metrics_mid}
\end{figure}

\paragraph{Duration distribution.}

\cref{fig:qualitative} shows the distribution of generated storyboard durations. In our experiments, the target duration is set to three to five minutes. A2Summ exhibits relatively broad distributions, reflecting variation in source-video length. In contrast, TeaserGen and REGen generally produce shorter videos because they are designed specifically for teaser generation. Both Clips2Story variants tend to exceed the target duration. Clips2Story-NF remains closer to the desired range, whereas Clips2Story-ND exceeds it by a larger margin because its planning process also accounts for the estimated duration of the generated narration. Under the finer-grained clip-level setting, the duration distributions of both variants shift closer to the target range, as shorter candidate clips provide the LLM with more precise control over the final storyboard duration.

\paragraph{Keyword alignment.}

As shown in \cref{tab:objective_combined}, Clips2Story achieves its largest gains in KR-Narration. Clips2Story-ND ranks highest in both settings, followed by Clips2Story-NF. Both variants also achieve the highest or tied-highest KR-Frame, while Clips2Story-ND obtains the highest KR-Caption. Compared with the input clip pool, both variants improve relevance across all modalities. \Cref{fig:keyword_relevance_timeline} further illustrates these results over time. For the documentary \textit{Black and Jewish America: An Interwoven History}, the target keywords exhibit distinct relevance profiles in the input video, while the Clips2Story outputs for ``Strange Fruit'' maintain higher narration relevance than the baseline outputs across much of their normalized timelines. Together, the results indicate that keyword-based retrieval effectively anchors the generated storyboards to the target theme.

\paragraph{Cross-genre robustness.}

\cref{fig:genre_metrics_mid} compares performance across five genres with different temporal structures and speech densities. Clips2Story-NF achieves the highest OPC in every genre, with particularly strong results for the transcript-dense lecture and news categories; corresponding word-density statistics are reported in \cref{tab:duration_stats}. Both Clips2Story variants also maintain relatively stable VTGHLS across genres, whereas REGen drops substantially for films, lectures, and vlogs despite comparable CLIPScores. Keyword alignment is more modality-dependent: Clips2Story-ND achieves the highest or near-highest KR-Narration across genres, while KR-Frame differences are small and KR-Caption gains are less consistent. Overall, the proposed methods generalize across genres, with the clearest advantages in narration-based relevance and temporal-order preservation.

\begin{table}[t]
\centering
\footnotesize
\setlength{\tabcolsep}{2pt}
\renewcommand{\arraystretch}{1.05}
\caption{Subjective evaluation results under the scene-level setting on a
Likert scale from 1 to 5. Higher values are better. All values are reported
with 95\% confidence intervals.}
\label{tab:subjective}
\vspace{1ex}
\resizebox{\linewidth}{!}{%
\begin{tabular}{lcccccccc}
\toprule
& \multicolumn{5}{c}{Generated videos}
& \multicolumn{3}{c}{Generated textual storyboards} \\
\cmidrule(lr){2-6}
\cmidrule(lr){7-9}

Model
& Coherence
& \shortstack{Keyword\\alignment}
& Realness
& \shortstack{Narration\\effectiveness}
& \shortstack{Clip-selection\\effectiveness}
& Organization
& \shortstack{Keyword\\informativeness}
& Engagingness \\
\midrule

REGen {\scriptsize\citep{xu2025regen}}
& $3.17 \pm 0.11$
& $3.11 \pm 0.16$
& $2.82 \pm 0.17$
& $3.27 \pm 0.15$
& $3.03 \pm 0.17$
& $\mathbf{4.07 \pm 0.19}$
& $3.69 \pm 0.23$
& $3.73 \pm 0.27$ \\

Clips2Story-NF
& $3.69 \pm 0.13$
& $3.52 \pm 0.15$
& $3.58 \pm 0.15$
& --
& $3.58 \pm 0.15$
& --
& --
& -- \\

Clips2Story-ND
& $\mathbf{4.43 \pm 0.14}$
& $\mathbf{4.42 \pm 0.13}$
& $\mathbf{4.28 \pm 0.15}$
& $\mathbf{4.29 \pm 0.14}$
& $\mathbf{4.17 \pm 0.16}$
& $\mathbf{4.07 \pm 0.19}$
& $\mathbf{4.09 \pm 0.18}$
& $\mathbf{4.07 \pm 0.21}$ \\
\bottomrule
\end{tabular}%
}
\end{table}

\begin{table*}[t]
    \centering
    \footnotesize
    \setlength{\tabcolsep}{3pt}
    \caption{
        Ablation results under the clip-level setting. All variants are
        defined relative to the full Clips2Story-NF pipeline. All values
        are reported with 95\% confidence intervals.
    }
    \label{tab:ablation}
    \vspace{1ex}
    \resizebox{\textwidth}{!}{%
        \begin{tabular}{lcccccccc}
            \toprule
            Model
            & MCD (sec)
            & REP (\%)
            & CLIPScore
            & VTGHLS
            & KR-Narration $\uparrow$
            & KR-Frame $\uparrow$
            & KR-Caption $\uparrow$
            & OPC \\
            
            \midrule

            Clips2Story-NF (full model)
            & $14.52 \pm 2.90$
            & $0.63 \pm 0.36$
            & $0.54 \pm 0.02$
            & $1.05 \pm 0.03$
            & $\mathbf{0.24 \pm 0.02}$
            & $\mathbf{0.19 \pm 0.01}$
            & $0.22 \pm 0.02$
            & $0.27 \pm 0.09$ \\

            ~~w/o visual descriptors
            & $15.95 \pm 3.20$
            & $0.63 \pm 0.38$
            & $0.49 \pm 0.04$
            & $0.97 \pm 0.08$
            & $0.23 \pm 0.02$
            & $0.17 \pm 0.01$
            & $0.20 \pm 0.02$
            & $0.38 \pm 0.06$ \\

            ~~w/o transcript
            & $11.39 \pm 2.38$
            & $0.71 \pm 0.32$
            & $0.52 \pm 0.03$
            & $1.04 \pm 0.07$
            & $0.17 \pm 0.02$
            & $0.18 \pm 0.01$
            & $\mathbf{0.23 \pm 0.02}$
            & $0.01 \pm 0.06$ \\

            ~~w/o keyword-based clip retrieval
            & $12.52 \pm 2.40$
            & $0.75 \pm 3.90$
            & $0.51 \pm 0.04$
            & $0.99 \pm 0.08$
            & $0.17 \pm 0.02$
            & $0.17 \pm 0.01$
            & $0.21 \pm 0.02$
            & $0.27 \pm 0.07$ \\

            ~~w/ video captioning
            & $15.60 \pm 3.14$
            & $0.78 \pm 0.42$
            & $0.51 \pm 0.04$
            & $1.00 \pm 0.08$
            & $0.23 \pm 0.02$
            & $0.19 \pm 0.01$
            & $0.22 \pm 0.02$
            & $0.28 \pm 0.08$ \\

            \cmidrule(lr){1-9}

            ~~w/ Qwen3-32B {\scriptsize\citep{yang2025qwen3}}
            & $9.57 \pm 1.72$
            & $10.23 \pm 3.21$
            & $0.49 \pm 0.04$
            & $0.98 \pm 0.08$
            & $0.20 \pm 0.02$
            & $0.17 \pm 0.01$
            & $0.20 \pm 0.02$
            & $0.18 \pm 0.08$ \\

            ~~w/ Qwen2.5-VL-7B-Instruct {\scriptsize\citep{bai2025qwen25vl}}
            & $17.72 \pm 4.66$
            & $5.54 \pm 2.16$
            & $0.47 \pm 0.05$
            & $0.92 \pm 0.09$
            & $0.23 \pm 0.03$
            & $0.18 \pm 0.01$
            & $0.20 \pm 0.02$
            & $0.19 \pm 0.09$ \\

            \bottomrule
        \end{tabular}%
    }
\end{table*}

\subsection{Subjective Evaluation Results}

\cref{tab:subjective} shows that both Clips2Story variants consistently outperform REGen across all evaluated criteria, with the narration-driven Clips2Story-ND outperforming the narration-free Clips2Story-NF.
Clips2Story-ND also achieves high coherence and keyword alignment scores,
showing that narration-driven planning significantly improves logical flow and semantic faithfulness. Even without voiceovers, Clips2Story-NF surpasses REGen in coherence and realness, confirming that our retrieval-first approach selects higher-quality visual sequences. Conversely, REGen achieves a significantly lower realness score than the Clips2Story variants,
yielding outputs that viewers perceive as artificial or disjointed. While Clips2Story-ND achieves parity with REGen in organization, it significantly outperforms the baseline in keyword informativeness and engagingness. These large
margins confirm that storyboard generation using a filtered clip pool and text-based descriptors can yield compelling, contextually anchored stories that the baseline struggles to achieve.

\subsection{Ablation Studies}

We conduct ablation studies on a 50-video subset under the clip-level setting using Clips2Story-NF. We evaluate the effects of removing visual descriptors or transcripts, replacing keyword-based retrieval with random retrieval, replacing middle-frame captioning with video captioning, and replacing GPT-5.2~\citep{openai2025gpt52} with either the open-weight Qwen3-32B \citep{yang2025qwen3} or the vision-language model Qwen2.5-VL \citep{bai2025qwen25vl}. For random retrieval, we shuffle the full clip pool and sample clips using the same duration budget as keyword-based retrieval to satisfy the planner's context limit. All results are reported with 95\% confidence intervals.

As shown in \cref{tab:ablation}, visual and textual descriptors serve complementary roles. Removing visual descriptors decreases CLIPScore from 0.54 to 0.49, VTGHLS from 1.05 to 0.97, and frame- and caption-based keyword relevance, but increases OPC from 0.27 to 0.38, suggesting that visual descriptors primarily support semantic grounding. In contrast, removing transcripts reduces KR-Narration from 0.24 to 0.17 and OPC to 0.01 while largely preserving CLIPScore and VTGHLS, indicating that transcripts support narrative and temporal ordering. Random retrieval reduces keyword relevance across all modalities without improving mean OPC, showing that the planner cannot fully compensate for a less relevant candidate pool. The video-captioning variant performs similarly to the full model, with only modest decreases in CLIPScore and VTGHLS, suggesting that middle-frame captioning efficiently represents the temporally consistent clips produced by our segmentation pipeline.

Finally, replacing GPT-5.2 with Qwen3-32B or Qwen2.5-VL lowers CLIPScore, VTGHLS, and OPC while increasing REP. Qwen3-32B yields the highest REP at 10.23\%, while Qwen2.5-VL retains comparable KR-Narration but shows weaker semantic alignment and temporal ordering. These results show that Clips2Story supports both open-weight and vision-language planners, although neither matches GPT-5.2 across the evaluated metrics. Taken together, transcripts support narrative ordering, visual descriptors strengthen semantic grounding, and keyword-based retrieval improves candidate relevance. Middle-frame captioning provides an efficient visual representation, while planner strength remains critical for coherent, non-repetitive storyboards.

\subsection{Computational Cost}

We evaluate inference latency on the full dataset of 100 videos using a single NVIDIA L40S GPU. Clips2Story-NF requires an average of 6.18 minutes per video, equivalent to 16.68\% of the input duration. Clips2Story-ND requires 7.96 minutes per video on average, including 1.77 minutes for narration--visual matching. Multimodal metadata is extracted only once for each input video and can be reused across different editing prompts. Overall, both variants complete inference in substantially less time than the duration of the input video, demonstrating the computational feasibility of the proposed training-free pipelines.

\section{Discussion \& Limitations}
\label{sec:discussions}

While Clips2Story demonstrates robust performance, it has several limitations. First, relying solely on text-based descriptors means that the system overlooks critical non-textual elements, such as fine-grained visual details and background music continuity. This limits overall fidelity and can result in visually or acoustically jarring transitions. Future work could leverage native multimodal foundation models to process raw audiovisual streams directly. Second, our hard-filtering retrieval strategy, designed to manage LLM context limits, may discard structurally useful but semantically tangential B-roll. A dynamic retrieval mechanism could enable more complex narrative arcs. Moreover, by optimizing for narrative coherence, the LLM risks recontextualizing factual footage into misleading storylines. This potential for contextual manipulation and misinformation is a critical ethical concern, particularly in sensitive domains such as news and documentaries.

\begin{figure*}[!htbp]
    \centering
    \includegraphics[width=\textwidth]{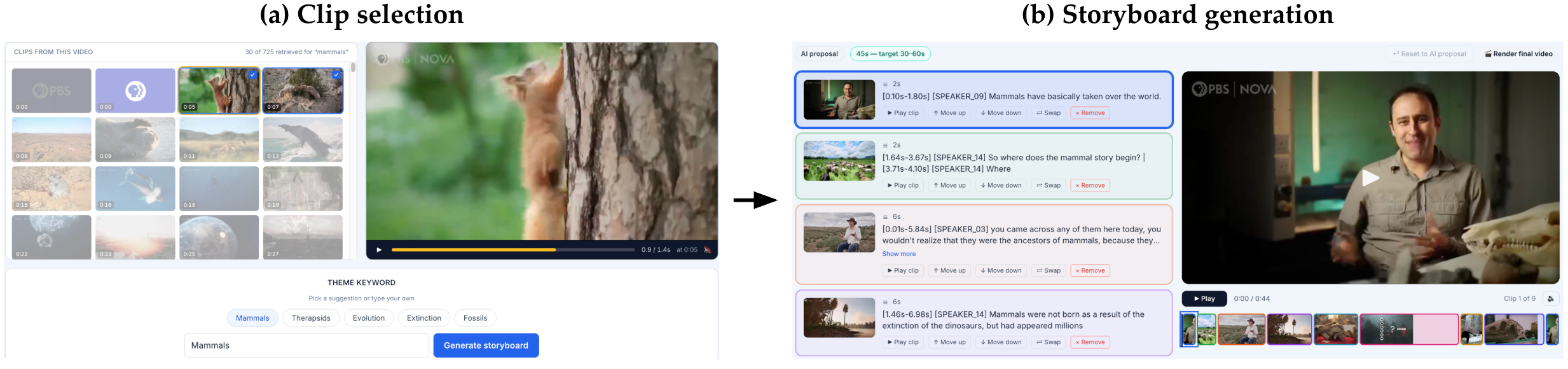}
    \caption{Prototype Clips2Story interface. (a) Given a user-input keyword, the system will suggest the most relevant video clips from the video library. The user may review and adjust the selection. (b) Given the selected video clips, the system will generate a storyboard with clip details and a live preview. The user may review, reorder, replace, and remove the video clips before the final rendering. A demo of the prototype can be found on our demo website.\cref{fn:demo}}
    \label{fig:editable_storyboard_workflow}
\end{figure*}

Finally, we show in \cref{fig:editable_storyboard_workflow} a prototype interface for the proposed Clips2Story system. The interface allows the user to enter a keyword that determines the theme of the video. The system will automatically suggest the most relevant clips from the video library for inclusion in the second stage, but the user can review and adjust the selection. The selected video clips will be passed to the second stage, where the system will suggest a storyboard with a live preview. The user can review, reorder, replace, or remove the video clips before the final rendering. The interface is designed to integrate directly into professional video editing workflows while preserving human agency and creativity. A demo of the prototype can be found on our demo website.\cref{fn:demo}



\section{Conclusion}
\label{sec:conclusion}

We introduced Clips2Story, a training-free, LLM-driven framework that generates coherent video storyboards from a non-chronologically ordered pool of video clips. To evaluate our approach, we developed a synthetic data generation pipeline that removes the original temporal order of the clips. We show that, given a pool of video clips and a user-specified keyword, the proposed system can effectively generate a coherent video storyboard by selecting and rearranging the input clips. Our human evaluation further shows that Clips2Story outperforms REGen in terms of coherence, keyword alignment, and realness.
Our framework highlights the potential of LLMs as active editorial agents, paving the way for more intelligent and scalable AI-assisted video editing tools.

\bibliographystyle{plainnat}
\bibliography{neurips}

@article{bai2025qwen25vl,
  title   = {{Qwen2.5-VL} Technical Report},
  author  = {Bai, Shuai and
             Chen, Keqin and
             Liu, Xuejing and
             Wang, Jialin and
             Ge, Wenbin and
             Song, Sibo and
             Dang, Kai and
             Wang, Peng and
             Wang, Shijie and
             Tang, Jun and
             Zhong, Humen and
             Zhu, Yuanzhi and
             Yang, Mingkun and
             Li, Zhaohai and
             Wan, Jianqiang and
             Wang, Pengfei and
             Ding, Wei and
             Fu, Zheren and
             Xu, Yiheng and
             Ye, Jiabo and
             Zhang, Xi and
             Xie, Tianbao and
             Cheng, Zesen and
             Zhang, Hang and
             Yang, Zhibo and
             Xu, Haiyang and
             Lin, Junyang},
  journal = {arXiv preprint arXiv:2502.13923},
  year    = {2025}
}

@article{yang2025qwen3,
  title={Qwen3 technical report},
  author={Yang, An and Li, Anfeng and Yang, Baosong and Zhang, Beichen and Hui, Binyuan and Zheng, Bo and Yu, Bowen and Gao, Chang and Huang, Chengen and Lv, Chenxu and others},
  journal={arXiv preprint arXiv:2505.09388},
  year={2025}
}

@article{wang2024qwen2,
  title={Qwen2-vl: Enhancing vision-language model's perception of the world at any resolution},
  author={Wang, Peng and Bai, Shuai and Tan, Sinan and Wang, Shijie and Fan, Zhihao and Bai, Jinze and Chen, Keqin and Liu, Xuejing and Wang, Jialin and Ge, Wenbin and others},
  journal={arXiv preprint arXiv:2409.12191},
  year={2024}
}

@article{kondratyuk2023videopoet,
  title={Videopoet: A large language model for zero-shot video generation},
  author={Kondratyuk, Dan and Yu, Lijun and Gu, Xiuye and Lezama, Jos{\'e} and Huang, Jonathan and Schindler, Grant and Hornung, Rachel and Birodkar, Vighnesh and Yan, Jimmy and Chiu, Ming-Chang and others},
  journal={arXiv preprint arXiv:2312.14125},
  year={2023}
}

@article{lian2023llm,
  title={Llm-grounded video diffusion models},
  author={Lian, Long and Shi, Baifeng and Yala, Adam and Darrell, Trevor and Li, Boyi},
  journal={arXiv preprint arXiv:2309.17444},
  year={2023}
}

@inproceedings{argaw2024scaling,
  title={Scaling up video summarization pretraining with large language models},
  author={Argaw, Dawit Mureja and Yoon, Seunghyun and Heilbron, Fabian Caba and Deilamsalehy, Hanieh and Bui, Trung and Wang, Zhaowen and Dernoncourt, Franck and Chung, Joon Son},
  booktitle={Proceedings of the IEEE/CVF Conference on Computer Vision and Pattern Recognition},
  pages={8332--8341},
  year={2024}
}

@inproceedings{singh2024previously,
  title={Previously on... from recaps to story summarization},
  author={Singh, Aditya Kumar and Srivastava, Dhruv and Tapaswi, Makarand},
  booktitle={Proceedings of the IEEE/CVF Conference on Computer Vision and Pattern Recognition},
  pages={13635--13646},
  year={2024}
}

@article{bain2023whisperx,
  title={Whisperx: Time-accurate speech transcription of long-form audio},
  author={Bain, Max and Huh, Jaesung and Han, Tengda and Zisserman, Andrew},
  journal={arXiv preprint arXiv:2303.00747},
  year={2023}
}

@inproceedings{liu2024grounding,
  title={Grounding dino: Marrying dino with grounded pre-training for open-set object detection},
  author={Liu, Shilong and Zeng, Zhaoyang and Ren, Tianhe and Li, Feng and Zhang, Hao and Yang, Jie and Jiang, Qing and Li, Chunyuan and Yang, Jianwei and Su, Hang and others},
  booktitle={European conference on computer vision},
  pages={38--55},
  year={2024},
  organization={Springer}
}

@inproceedings{kim2021conditional,
  title={Conditional variational autoencoder with adversarial learning for end-to-end text-to-speech},
  author={Kim, Jaehyeon and Kong, Jungil and Son, Juhee},
  booktitle={International conference on machine learning},
  pages={5530--5540},
  year={2021},
  organization={PMLR}
}

@inproceedings{soucek2024transnet,
  title={Transnet v2: An effective deep network architecture for fast shot transition detection},
  author={Soucek, Tom{\'a}s and Lokoc, Jakub},
  booktitle={Proceedings of the 32nd ACM International Conference on Multimedia},
  pages={11218--11221},
  year={2024}
}

@inproceedings{radford2021learning,
  title={Learning transferable visual models from natural language supervision},
  author={Radford, Alec and Kim, Jong Wook and Hallacy, Chris and Ramesh, Aditya and Goh, Gabriel and Agarwal, Sandhini and Sastry, Girish and Askell, Amanda and Mishkin, Pamela and Clark, Jack and others},
  booktitle={International conference on machine learning},
  pages={8748--8763},
  year={2021},
  organization={PmLR}
}

@inproceedings{li2022blip,
  title={Blip: Bootstrapping language-image pre-training for unified vision-language understanding and generation},
  author={Li, Junnan and Li, Dongxu and Xiong, Caiming and Hoi, Steven},
  booktitle={International conference on machine learning},
  pages={12888--12900},
  year={2022},
  organization={PMLR}
}

@article{kong2020panns,
  title={Panns: Large-scale pretrained audio neural networks for audio pattern recognition},
  author={Kong, Qiuqiang and Cao, Yin and Iqbal, Turab and Wang, Yuxuan and Wang, Wenwu and Plumbley, Mark D},
  journal={IEEE/ACM Transactions on Audio, Speech, and Language Processing},
  volume={28},
  pages={2880--2894},
  year={2020},
  publisher={IEEE}
}

@inproceedings{reimers2019sentence,
  title={Sentence-bert: Sentence embeddings using siamese bert-networks},
  author={Reimers, Nils and Gurevych, Iryna},
  booktitle={Proceedings of the 2019 conference on empirical methods in natural language processing and the 9th international joint conference on natural language processing (EMNLP-IJCNLP)},
  pages={3982--3992},
  year={2019}
}

@misc{openai2025gpt52,
  author       = {{OpenAI}},
  title        = {Introducing {GPT}-5.2},
  year         = {2025},
  month        = dec,
  howpublished = {\url{https://openai.com/index/introducing-gpt-5-2/}}
}

@article{liu2024lost,
  title={Lost in the middle: How language models use long contexts},
  author={Liu, Nelson F and Lin, Kevin and Hewitt, John and Paranjape, Ashwin and Bevilacqua, Michele and Petroni, Fabio and Liang, Percy},
  journal={Transactions of the association for computational linguistics},
  volume={12},
  pages={157--173},
  year={2024}
}

@article{lin2023videoxum,
  title={Videoxum: Cross-modal visual and textural summarization of videos},
  author={Lin, Jingyang and Hua, Hang and Chen, Ming and Li, Yikang and Hsiao, Jenhao and Ho, Chiuman and Luo, Jiebo},
  journal={IEEE Transactions on Multimedia},
  volume={26},
  pages={5548--5560},
  year={2023},
  publisher={IEEE}
}

@inproceedings{he2023align,
  title={Align and attend: Multimodal summarization with dual contrastive losses},
  author={He, Bo and Wang, Jun and Qiu, Jielin and Bui, Trung and Shrivastava, Abhinav and Wang, Zhaowen},
  booktitle={Proceedings of the IEEE/CVF conference on computer vision and pattern recognition},
  pages={14867--14878},
  year={2023}
}

@inproceedings{bose2023movieclip,
  title={Movieclip: Visual scene recognition in movies},
  author={Bose, Digbalay and Hebbar, Rajat and Somandepalli, Krishna and Zhang, Haoyang and Cui, Yin and Cole-McLaughlin, Kree and Wang, Huisheng and Narayanan, Shrikanth},
  booktitle={Proceedings of the IEEE/CVF Winter Conference on Applications of Computer Vision},
  pages={2083--2092},
  year={2023}
}

@article{xu2024teasergen,
  title={Teasergen: Generating teasers for long documentaries},
  author={Xu, Weihan and Liang, Paul Pu and Kim, Haven and McAuley, Julian and Berg-Kirkpatrick, Taylor and Dong, Hao-Wen},
  journal={arXiv preprint arXiv:2410.05586},
  year={2024}
}

@article{xu2025regen,
  title={REGen: Multimodal Retrieval-Embedded Generation for Long-to-Short Video Editing},
  author={Xu, Weihan and Ma, Yimeng and Huang, Jingyue and Li, Yang and Ma, Wenye and Berg-Kirkpatrick, Taylor and McAuley, Julian and Liang, Paul Pu and Dong, Hao-Wen},
  journal={arXiv preprint arXiv:2505.18880},
  year={2025}
}

@article{lin2023videodirectorgpt,
  title={Videodirectorgpt: Consistent multi-scene video generation via llm-guided planning},
  author={Lin, Han and Zala, Abhay and Cho, Jaemin and Bansal, Mohit},
  journal={arXiv preprint arXiv:2309.15091},
  year={2023}
}

@article{yeh2026vidmento,
  title={Vidmento: Creating Video Stories Through Context-Aware Expansion With Generative Video},
  author={Yeh, Catherine and Truong, Anh and Dontcheva, Mira and Wang, Bryan},
  journal={arXiv preprint arXiv:2601.22013},
  year={2026}
}

@inproceedings{wang2024lave,
  title={Lave: Llm-powered agent assistance and language augmentation for video editing},
  author={Wang, Bryan and Li, Yuliang and Lv, Zhaoyang and Xia, Haijun and Xu, Yan and Sodhi, Raj},
  booktitle={Proceedings of the 29th International Conference on Intelligent User Interfaces},
  pages={699--714},
  year={2024}
}


\clearpage

\appendix

\crefalias{section}{appendix}
\setcounter{secnumdepth}{1} 

\section{Comparisons to Prior Work}
\label{app:comparisons}

\cref{tab:comparison} compares Clips2Story with related methods.

\begin{table*}[!htbp]
\centering
\small
\setlength{\tabcolsep}{3pt}
\caption{Comparison of Clips2Story with related methods. Unlike existing baselines, our approach operates on an unordered pool of video clips and supports keyword-guided clip retrieval, narrative ordering, and connective narration. Here, Summ. = summarization, Edit. = editing, Gen. = generation, Ord. = ordered, Unord. = unordered, retr. = retrieval, Narr. = narrative, Conn. = connective, and narr. = narration. $^{*}$Achieved via our Clips2Story-ND pipeline.}
\vspace{1ex}
\begin{tabular}{lcccccccc}
\toprule
& \multicolumn{2}{c}{{Task type}} & \multicolumn{3}{c}{{Inputs}} & \multicolumn{3}{c}{{Capabilities}} \\
\cmidrule(lr){2-3} \cmidrule(lr){4-6} \cmidrule(lr){7-9}
{Model} 
& \shortstack{{Summ.} \\ {/ Edit.}} 
& {Gen.} 
& {Ord.} 
& {Unord.} 
& \shortstack{{Keyword} \\ {guided}} 
& \shortstack{{Clip} \\ {retr.}} 
& \shortstack{{Narr.} \\ {ord.}} 
& \shortstack{{Conn.} \\ {narr.}} \\
\midrule
MovieClip \citep{bose2023movieclip} & \yes & \no & \yes & \no & \no & \no & \no & \no \\
A2Summ \citep{he2023align} & \yes & \no & \yes & \no & \no & \yes & \no & \no \\
LfVS \citep{argaw2024scaling} & \yes & \no & \yes & \no & \no & \yes & \no & \no \\
TaleSumm \citep{singh2024previously} & \yes & \no & \yes & \no & \no & \yes & \no & \no \\
VideoXUM \citep{lin2023videoxum} & \yes & \no & \yes & \no & \no & \no & \no & \yes \\
LAVE \citep{wang2024lave} & \yes & \no & \yes & \no & \yes & \yes & \yes & \no \\
TeaserGen \citep{xu2024teasergen} & \yes & \no & \yes & \no & \no & \yes & \yes & \yes \\
REGen \citep{xu2025regen} & \yes & \no & \yes & \no & \no & \yes & \yes & \yes \\
LLM-grounded \citep{lian2023llm} & \no & \yes & \no & \no & \yes & \no & \yes & \no \\
Vidmento \citep{yeh2026vidmento} & \yes & \yes & \yes & \no & \yes & \no & \yes & \no \\
VideoPoet \citep{kondratyuk2023videopoet} & \yes & \yes & \yes & \no & \yes & \no & \yes & \no \\
\midrule
{Clips2Story (ours)} & \textbf{\yes} & \textbf{\no} & \textbf{\no} & \textbf{\yes} & \textbf{\yes} & \textbf{\yes} & \textbf{\yes} & \textbf{\yes}$^{*}$ \\
\bottomrule
\end{tabular}%
\label{tab:comparison}
\end{table*}




\section{Prompt Template for the Narration-Free Variant}
\label{app:nf_prompt}

The prompt used for the Narration-Free pipeline is provided below. In practice, the \texttt{AVAILABLE CLIPS (UNORDERED POOL)} field is populated dynamically with the retrieved candidate clips for each example.

\begin{quote}
\small
\textbf{System / Instruction Prompt}

You are an expert AI video editor and planning assistant. Convert the user's editing request into a storyboard (sequence of footage segments) in strict JSON.

\vspace{0.5em}
\textbf{USER REQUEST}

Create a catchy 3--5 minute mini-story about \texttt{\{keyword\}}. Open with a strong hook, build curiosity and progression, and end with a satisfying payoff. Select and sequence footage that naturally builds this narrative.

\vspace{0.5em}
\textbf{AVAILABLE CLIPS (UNORDERED POOL)}

Each clip is independent raw footage. There is no implied chronological relationship by list order. Use only the provided \texttt{clip\_id} strings when referencing clips in your timeline. Each clip includes transcript, caption, background, and entities.

\begin{verbatim}
- clip_id={clip_id_1}, duration={duration_1}
  transcript="{transcript_1}"
  caption="{caption_1}"
  background="{background_1}"
  entities="{entities_1}"

- clip_id={clip_id_2}, duration={duration_2}
  transcript="{transcript_2}"
  caption="{caption_2}"
  background="{background_2}"
  entities="{entities_2}"

...
\end{verbatim}

\vspace{0.5em}
\textbf{TASK}

Produce an ordered sequence of footage segments. Each item in the sequence has a \texttt{"segments"} list containing the footage segments that tell the story.

You may select, skip, trim, or reorder clips as needed. When trimming, define the exact \texttt{start\_time} and \texttt{end\_time} to extract the specific part of the clip you want to use, while respecting transcript sentence boundaries. Choose as many sequence items as needed so that the total runtime is approximately 3--5 minutes.

\vspace{0.5em}
\textbf{RULES}
\begin{enumerate}
    \item Compose a coherent story arc with a strong hook, clear progression, and a satisfying payoff. Infer intent from the user request, including topic, tone, and pacing.
    \item Select footage using transcript, caption, background, and entity evidence; prefer strong matches. Ensure that the sequence of clips flows naturally to tell the story.
    \item Choose clips and precisely define their \texttt{start\_time}/\texttt{end\_time} to trim the footage to the exact moments needed. Do not trim mid-sentence. If a clip contains speech, the \texttt{start\_time} and \texttt{end\_time} must exactly match transcript timestamp brackets, never intermediate points. Do not feel constrained to use the full clip duration.
    \item Choose as many segments as needed so that the total runtime, defined as the sum of all selected clip durations, is approximately 3--5 minutes.
    \item Do not invent any clips. Do not invent or output timestamps outside the bounds of the original clip duration.
    \item For each selected segment, include \texttt{"duration"} (computed as \texttt{end\_time - start\_time}) and \texttt{"transcript"} containing only the transcript text that falls within the selected time span, preserving speaker labels and relative timestamps when present.
    \begin{itemize}
        \item Do not paraphrase, summarize, or invent transcript text.
        \item If no transcript is available for that clip or trimmed segment, output \texttt{"transcript": ""}.
    \end{itemize}
\end{enumerate}

\vspace{0.5em}
\textbf{HARD CONSTRAINTS}
\begin{itemize}
    \item Each segment's \texttt{clip\_id} must exactly match an available clip.
    \item Do not invent clips.
    \item Output must be valid JSON only, with no markdown or extra text.
\end{itemize}

\vspace{0.5em}
\textbf{OUTPUT SCHEMA (STRICT)}
\begin{verbatim}
[
  {
    "segments": [
      {
        "clip_id": "c_91f2a3",
        "start_time": 0.0,
        "end_time": 0.0,
        "duration": 0.0,
        "transcript": ""
      }
    ]
  }
]
\end{verbatim}

Output only the JSON array, with no markdown and no extra text.
\end{quote}

\section{Prompt Template for the Narration-Driven Variant}
\label{app:nd_prompt}

The prompt used for the Narration-Driven pipeline is provided below. In practice, the \texttt{AVAILABLE CLIPS (UNORDERED POOL)} field is populated dynamically with the retrieved candidate clips for each example.

\begin{quote}
\small
\textbf{System / Instruction Prompt}

You are an expert AI video editor and planning assistant. Convert the user's editing request into a storyboard (narration + supporting footage segments) in strict JSON.

\vspace{0.5em}
\textbf{USER REQUEST}

Create a catchy 3--5 minute mini-story about \texttt{\{keyword\}}. Open with a strong hook, build curiosity and progression, and end with a satisfying payoff. Write short narrations that guide the viewer, and for each narration select footage that will visually support what the narration just set up.

\vspace{0.5em}
\textbf{CRITICAL -- RENDERING ORDER}
\begin{itemize}
    \item Narration is played first; then the selected footage segments for that narration play. Narration is not played over footage.
    \item Therefore, write each narration as a lead-in for the clips that follow it. The footage should visually support what the narration just set up using transcript, caption, background, and entity evidence when available.
\end{itemize}

\vspace{0.5em}
\textbf{AVAILABLE CLIPS (UNORDERED POOL)}

Each clip is independent raw footage. There is no implied chronological relationship by list order. Use only the provided \texttt{clip\_id} strings when referencing clips in your timeline. Each clip includes transcript, caption, background, and entities.

\begin{verbatim}
- clip_id={clip_id_1}, duration={duration_1}
  transcript="{transcript_1}"
  caption="{caption_1}"
  background="{background_1}"
  entities="{entities_1}"

- clip_id={clip_id_2}, duration={duration_2}
  transcript="{transcript_2}"
  caption="{caption_2}"
  background="{background_2}"
  entities="{entities_2}"

...
\end{verbatim}

\vspace{0.5em}
\textbf{TASK}

Produce a list of sequence items. Each item in the sequence has:
(1) \texttt{"narration"}: short text for that moment (no timestamps; narration duration will be determined later by TTS), and
(2) \texttt{"segments"}: footage segments that play after this narration and visually support it.

You may select, skip, trim, or reorder clips as needed. When trimming, define the exact \texttt{start\_time} and \texttt{end\_time} to extract the specific part of the clip you want to use (respecting transcript sentence boundaries). Choose as many sequence items as needed so that the total runtime (narration + clip durations) is approximately 3--5 minutes.

\vspace{0.5em}
\textbf{RULES}
\begin{enumerate}
    \item Compose a coherent story arc with a strong hook, clear progression, and a satisfying payoff. Infer intent from the user request, including topic, tone, and pacing.
    \item Select footage using transcript, caption, background, and entity evidence; prefer strong matches. Each narration's segments should visually support what the narration just set up.
    \item Choose clips and precisely define their \texttt{start\_time}/\texttt{end\_time} to trim the footage to the exact moments needed. Do not trim mid-sentence. If a clip contains speech, the \texttt{start\_time} and \texttt{end\_time} must exactly match transcript timestamp brackets, never intermediate points. Do not feel constrained to use the full clip duration.
    \item Choose as many sequence items as needed so that the total runtime (TTS narration duration + all selected clip durations) is approximately 3--5 minutes. Do not add timestamps for narration; only provide the narration text.
    \item Do not invent any clips. Do not invent or output timestamps outside the bounds of the original clip duration.
    \item For each selected segment, include \texttt{"duration"} (computed as \texttt{end\_time - start\_time}) and \texttt{"transcript"} containing only the transcript text that falls within the selected time span, preserving speaker labels and relative timestamps when present.
    \begin{itemize}
        \item Do not paraphrase, summarize, or invent transcript text.
        \item If no transcript is available for that clip or trimmed segment, output \texttt{"transcript": ""}.
    \end{itemize}
\end{enumerate}

\vspace{0.5em}
\textbf{HARD CONSTRAINTS}
\begin{itemize}
    \item Each segment's \texttt{clip\_id} must exactly match an available clip.
    \item Do not invent clips.
    \item Output must be valid JSON only, with no markdown or extra text.
\end{itemize}

\vspace{0.5em}
\textbf{OUTPUT SCHEMA (STRICT)}
\begin{verbatim}
[
  {
    "narration": "string",
    "segments": [
      {
        "clip_id": "c_91f2a3",
        "start_time": 0.0,
        "end_time": 0.0,
        "duration": 0.0,
        "transcript": ""
      }
    ]
  }
]
\end{verbatim}

Output only the JSON array, with no markdown and no extra text.
\end{quote}

\section{Dataset Construction Details}
\label{app:dataset_details}

Below, we provide the technical specifications and hyperparameters used to process the source videos into the clip-level and scene-level candidate pools.

\paragraph{Clip-level extraction.}

In the clip-level setting, initial shots are detected using TransNetV2 \citep{soucek2024transnet}, generating the finest level of temporal granularity used in our experiments.

\paragraph{Scene-level grouping.}

In the scene-level setting, similarity is computed only between neighboring shots using multimodal features: visual similarity from CLIP ViT-B/32 \citep{radford2021learning} applied to the middle frame of each shot, caption similarity from CLIP ViT-B/32 applied to BLIP-generated captions \citep{li2022blip}, audio similarity from PANNs \citep{kong2020panns}, and transcript similarity from Sentence-Transformer embeddings \citep{reimers2019sentence} of ASR text. These four signals are combined with weights 0.30 (visual), 0.25 (caption), 0.25 (audio), and 0.20 (ASR) to produce a single similarity score for each adjacent pair.

For each cut, we define a boundary value as $1 - \mathrm{similarity}$, where a higher boundary value indicates a stronger semantic or perceptual change between adjacent shots. To reduce noisy segmentation, the boundary sequence is smoothed with a median filter. For each video, a threshold $T$ is set as the 75th percentile of that video’s boundary distribution. Neighboring shots are merged into the same group while their boundary value is less than or equal to $T$; when the boundary exceeds $T$, a new group is started. To prevent excessively long grouped units, we impose a maximum group duration of 3 minutes.

\section{Detailed Evaluation Metric Formulations}
\label{app:eval_metrics}

We provide the mathematical definitions and technical implementation
details for the evaluation metrics introduced in the Experimental Setup section of the main paper.

\paragraph{Order preservation coefficient (OPC).}

Let $O=(o_1,\dots,o_n)$ be the deduplicated sequence of selected shots in the outputs and let $p(o_i)$ be the original position of shot $o_i$ in the source video. Excluding ties, we compute OPC as:

\begin{equation}
\mathrm{OPC} =
\begin{cases}
1, & n \le 1 \text{ or } P=0,\\[4pt]
\dfrac{C-D}{P}, & \text{otherwise},
\end{cases}
\end{equation}

where $C$ and $D$ are the numbers of concordant and discordant shot pairs, and $P=C+D$.

\paragraph{Keyword relevance (KR).}

We measure keyword relevance separately for narration/transcripts, video frames, and visual captions. Each modality is evaluated over its own set of approximately uniform textual or temporal units. The corresponding KR score is the unweighted mean cosine similarity between the target keyword and these units.

\textit{Narration and transcript relevance.}
For each segment in the generated timeline, we use the generated narration for narration-driven segments and the ASR transcript of the selected source clip otherwise. We concatenate the resulting text in timeline order and divide it at sentence boundaries into chunks of approximately 50 words. Let $\mathbf{k}^{\mathrm{st}}$ denote the sentence-transformer embedding of the target keyword \citep{reimers2019sentence}, and let $\mathbf{d}_j$ denote the embedding of the $j$-th text chunk. Given $J$ chunks, spoken-text relevance is defined as

\begin{equation}
\mathrm{KR}_{\mathrm{Narration}}
=
\frac{1}{J}
\sum_{j=1}^{J}
\cos\left(
\mathbf{k}^{\mathrm{st}},
\mathbf{d}_j
\right).
\end{equation}

\textit{Frame relevance.}
We uniformly sample the output timeline at $0.1$ FPS. Let $\mathbf{k}^{\mathrm{clip}}$ denote the OpenCLIP ViT-B/32 embedding of the target keyword \citep{radford2021learning}, and let $\mathbf{f}_m$ denote the embedding of the $m$-th sampled frame. Given $M$ sampled frames, frame relevance is computed as

\begin{equation}
\mathrm{KR}_{\mathrm{Frame}}
=
\frac{1}{M}
\sum_{m=1}^{M}
\cos\left(
\mathbf{k}^{\mathrm{clip}},
\mathbf{f}_m
\right).
\end{equation}

\textit{Caption relevance.}
At each frame-sampling timestamp, we caption a 10-second source-video window centered on that timestamp using Qwen2-VL-7B-Instruct \citep{wang2024qwen2}. Let $\mathbf{c}_m$ denote the OpenCLIP text embedding of the caption associated with the $m$-th sampled window. Caption relevance is computed as

\begin{equation}
\mathrm{KR}_{\mathrm{Caption}}
=
\frac{1}{M}
\sum_{m=1}^{M}
\cos\left(
\mathbf{k}^{\mathrm{clip}},
\mathbf{c}_m
\right).
\end{equation}

\section{Subjective Evaluation Questions for Clips-to-storyboard Generation}
\label{app:subjective_video_questions}

In our subjective evaluation survey, we ask the following questions to evaluate our proposed pipelines for storyboard generation.

\begin{itemize}
    \item \textbf{Coherence:} To what extent does the video maintain coherence and a smooth flow, such that each segment transitions logically and the overall viewing experience feels seamless?
    
    \item \textbf{Keyword alignment:} To what extent does the video align with the target keyword?
    
    \item \textbf{Realness:} To what extent does the video appear realistic and convincing overall?
    
    \item \textbf{Narration effectiveness:} To what extent do the narrations help you understand the connections between clips and guide your understanding of how the video progresses?
    
    \item \textbf{Selected clips effectiveness:} To what extent do the selected video clips provide useful and informative content for the video?
\end{itemize}

\section{Subjective Evaluation Questions for Generated Textual Storyboards}
\label{app:subjective_narration_questions}

In our subjective evaluation survey, we ask the following questions to evaluate our proposed pipelines for narration generation.

\begin{itemize}
    \item \textbf{Organization:} To what extent does the narration present a well-organized and consistent storyline?
    
    \item \textbf{Keyword informativeness:} To what extent does the narration clearly introduce the main characters or events related to the target keyword?
    
    \item \textbf{Engagingness:} To what extent does the narration keep you engaged and build curiosity about what happens next?
\end{itemize}

\section{Additional Quantitative Results}
\label{app:quant_results}

In this section, we provide a more granular breakdown of our model's performance. Specifically, we analyze the results across five distinct video genres under both scene-level and clip-level settings.

\begin{figure}[t]
    \centering
    \includegraphics[width=\columnwidth]{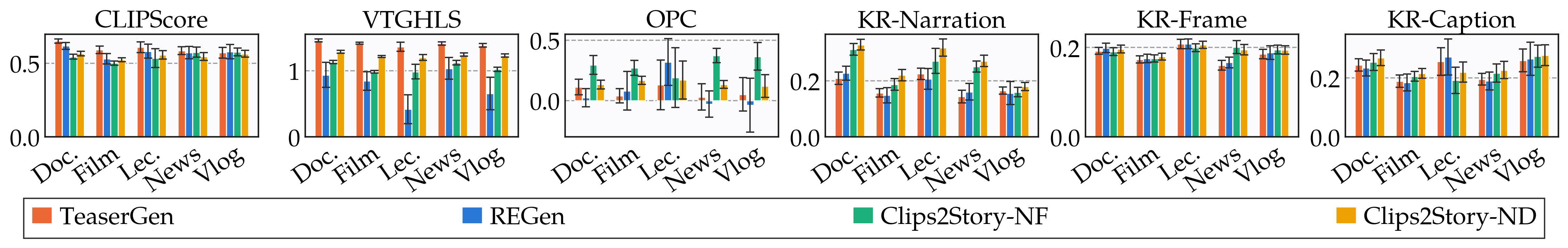}
    \caption{Genre-wise comparison of CLIPScore, VTGHLS, OPC, KR-Narration, KR-Frame, and KR-Caption under the clip-level setting. Error bars indicate 95\% confidence intervals. Specific numbers can be found in  \cref{tab:cross_genre_clip_level}.}
    \label{fig:duration_metrics_combined_clip}
\end{figure}


\FloatBarrier

\begin{table*}[!htbp]
\centering
\small
\setlength{\tabcolsep}{3pt}
\caption{Performance of A2Summ, TeaserGen, REGen, Clips2Story-NF, and Clips2Story-ND across five video genres under the scene-level setting, compared with the input video clips. Values are reported with 95\% confidence intervals.}
\label{tab:cross_genre_scene_level}
\vspace{1ex}
\resizebox{\textwidth}{!}{%
\begin{tabular}{lcccccccc}
\toprule
Model
& MCD (sec)
& REP (\%)
& CLIPScore
& VTGHLS
& OPC
& KR-Narration $\uparrow$
& KR-Frame $\uparrow$
& KR-Caption $\uparrow$ \\
\midrule

\multicolumn{9}{c}{\textbf{Documentary}} \\
\midrule
Input video clips
& $4.13 \pm 0.39$
& $0.00 \pm 0.00$
& $0.49 \pm 0.01$
& $1.02 \pm 0.03$
& $1.00 \pm 0.00$
& $0.17 \pm 0.02$
& $0.17 \pm 0.01$
& $0.23 \pm 0.02$ \\
\cmidrule(lr){1-9}
A2Summ
& $3.99 \pm 0.38$
& $0.00 \pm 0.00$
& $0.50 \pm 0.01$
& $1.07 \pm 0.02$
& $0.05 \pm 0.04$
& $0.19 \pm 0.02$
& $0.17 \pm 0.01$
& $0.23 \pm 0.02$ \\
TeaserGen \citep{xu2024teasergen}
& $3.48 \pm 0.10$
& $19.99 \pm 3.28$
& $0.67 \pm 0.02$
& $1.48 \pm 0.02$
& $0.04 \pm 0.07$
& $0.21 \pm 0.02$
& $0.19 \pm 0.01$
& $0.24 \pm 0.02$ \\
REGen \citep{xu2025regen}
& $3.56 \pm 0.41$
& $20.50 \pm 12.76$
& $0.59 \pm 0.04$
& $1.00 \pm 0.12$
& $-0.07 \pm 0.12$
& $0.22 \pm 0.03$
& $0.19 \pm 0.01$
& $0.23 \pm 0.03$ \\
\cmidrule(lr){1-9}
Clips2Story-NF
& $19.03 \pm 2.29$
& $0.29 \pm 0.29$
& $0.57 \pm 0.02$
& $1.15 \pm 0.03$
& $0.48 \pm 0.07$
& $0.27 \pm 0.02$
& $0.19 \pm 0.01$
& $\mathbf{0.26 \pm 0.03}$ \\
Clips2Story-ND
& $7.78 \pm 0.47$
& $10.07 \pm 1.08$
& $0.58 \pm 0.02$
& $1.24 \pm 0.03$
& $0.31 \pm 0.05$
& $\mathbf{0.30 \pm 0.02}$
& $\mathbf{0.20 \pm 0.01}$
& $\mathbf{0.26 \pm 0.03}$ \\
\midrule

\multicolumn{9}{c}{\textbf{Film}} \\
\midrule
Input video clips
& $9.66 \pm 1.83$
& $0.00 \pm 0.00$
& $0.47 \pm 0.00$
& $0.90 \pm 0.02$
& $1.00 \pm 0.00$
& $0.11 \pm 0.01$
& $0.16 \pm 0.01$
& $0.20 \pm 0.02$ \\
\cmidrule(lr){1-9}
A2Summ
& $9.18 \pm 1.60$
& $0.00 \pm 0.00$
& $0.49 \pm 0.01$
& $0.96 \pm 0.01$
& $0.10 \pm 0.09$
& $0.12 \pm 0.01$
& $0.16 \pm 0.01$
& $0.20 \pm 0.01$ \\
TeaserGen \citep{xu2024teasergen}
& $4.08 \pm 0.19$
& $8.43 \pm 2.25$
& $0.59 \pm 0.02$
& $1.43 \pm 0.01$
& $0.01 \pm 0.06$
& $0.16 \pm 0.02$
& $0.17 \pm 0.01$
& $0.18 \pm 0.02$ \\
REGen \citep{xu2025regen}
& $4.61 \pm 0.57$
& $10.55 \pm 9.08$
& $0.55 \pm 0.04$
& $0.66 \pm 0.22$
& $-0.06 \pm 0.20$
& $0.18 \pm 0.02$
& $0.17 \pm 0.01$
& $0.19 \pm 0.04$ \\
\cmidrule(lr){1-9}
Clips2Story-NF
& $27.14 \pm 4.87$
& $0.15 \pm 0.10$
& $0.51 \pm 0.02$
& $0.99 \pm 0.03$
& $0.40 \pm 0.07$
& $0.17 \pm 0.02$
& $\mathbf{0.18 \pm 0.01}$
& $0.20 \pm 0.02$ \\
Clips2Story-ND
& $9.93 \pm 0.84$
& $6.33 \pm 0.69$
& $0.51 \pm 0.02$
& $1.15 \pm 0.02$
& $0.28 \pm 0.05$
& $\mathbf{0.20 \pm 0.02}$
& $\mathbf{0.18 \pm 0.01}$
& $\mathbf{0.22 \pm 0.02}$ \\
\midrule

\multicolumn{9}{c}{\textbf{Lecture}} \\
\midrule
Input video clips
& $165.22 \pm 129.49$
& $0.01 \pm 0.01$
& $0.50 \pm 0.02$
& $1.02 \pm 0.02$
& $1.00 \pm 0.00$
& $0.21 \pm 0.02$
& $0.19 \pm 0.01$
& $0.15 \pm 0.02$ \\
\cmidrule(lr){1-9}
A2Summ
& $28.21 \pm 15.19$
& $0.00 \pm 0.00$
& $0.55 \pm 0.03$
& $1.05 \pm 0.05$
& $0.33 \pm 0.16$
& $0.23 \pm 0.02$
& $0.19 \pm 0.01$
& $0.18 \pm 0.03$ \\
TeaserGen \citep{xu2024teasergen}
& $4.82 \pm 0.45$
& $29.31 \pm 6.42$
& $0.60 \pm 0.03$
& $1.35 \pm 0.07$
& $0.16 \pm 0.18$
& $0.22 \pm 0.02$
& $\mathbf{0.21 \pm 0.01}$
& $0.25 \pm 0.04$ \\
REGen \citep{xu2025regen}
& $8.23 \pm 3.95$
& $14.74 \pm 13.58$
& $0.59 \pm 0.04$
& $0.55 \pm 0.23$
& $0.27 \pm 0.29$
& $0.22 \pm 0.03$
& $\mathbf{0.21 \pm 0.01}$
& $\mathbf{0.26 \pm 0.05}$ \\
\cmidrule(lr){1-9}
Clips2Story-NF
& $41.54 \pm 8.15$
& $0.47 \pm 0.26$
& $0.52 \pm 0.06$
& $0.98 \pm 0.11$
& $0.71 \pm 0.09$
& $0.26 \pm 0.04$
& $0.20 \pm 0.01$
& $0.19 \pm 0.04$ \\
Clips2Story-ND
& $14.17 \pm 2.63$
& $17.85 \pm 8.63$
& $0.56 \pm 0.03$
& $1.18 \pm 0.03$
& $0.32 \pm 0.13$
& $\mathbf{0.30 \pm 0.03}$
& $\mathbf{0.21 \pm 0.01}$
& $0.22 \pm 0.04$ \\
\midrule

\multicolumn{9}{c}{\textbf{News}} \\
\midrule
Input video clips
& $8.44 \pm 1.96$
& $0.05 \pm 0.01$
& $0.53 \pm 0.03$
& $1.05 \pm 0.03$
& $1.00 \pm 0.00$
& $0.13 \pm 0.02$
& $0.15 \pm 0.01$
& $0.17 \pm 0.02$ \\
\cmidrule(lr){1-9}
A2Summ
& $8.51 \pm 1.10$
& $0.00 \pm 0.00$
& $0.58 \pm 0.04$
& $1.13 \pm 0.04$
& $0.21 \pm 0.09$
& $0.14 \pm 0.02$
& $0.15 \pm 0.01$
& $0.18 \pm 0.02$ \\
TeaserGen \citep{xu2024teasergen}
& $4.26 \pm 0.28$
& $19.38 \pm 4.77$
& $0.60 \pm 0.03$
& $1.41 \pm 0.03$
& $-0.02 \pm 0.11$
& $0.15 \pm 0.02$
& $0.16 \pm 0.01$
& $0.18 \pm 0.02$ \\
REGen \citep{xu2025regen}
& $4.48 \pm 1.12$
& $16.20 \pm 12.32$
& $0.62 \pm 0.05$
& $0.79 \pm 0.24$
& $0.13 \pm 0.24$
& $0.14 \pm 0.03$
& $0.17 \pm 0.01$
& $0.20 \pm 0.03$ \\
\cmidrule(lr){1-9}
Clips2Story-NF
& $24.87 \pm 4.35$
& $3.20 \pm 1.99$
& $0.62 \pm 0.04$
& $1.14 \pm 0.04$
& $0.61 \pm 0.09$
& $0.24 \pm 0.02$
& $\mathbf{0.20 \pm 0.01}$
& $0.22 \pm 0.03$ \\
Clips2Story-ND
& $8.98 \pm 1.33$
& $20.51 \pm 3.70$
& $0.59 \pm 0.03$
& $1.23 \pm 0.03$
& $0.26 \pm 0.05$
& $\mathbf{0.26 \pm 0.02}$
& $\mathbf{0.20 \pm 0.01}$
& $\mathbf{0.23 \pm 0.03}$ \\
\midrule

\multicolumn{9}{c}{\textbf{Vlog}} \\
\midrule
Input video clips
& $74.36 \pm 62.64$
& $0.02 \pm 0.01$
& $0.53 \pm 0.02$
& $0.99 \pm 0.03$
& $1.00 \pm 0.00$
& $0.13 \pm 0.02$
& $0.18 \pm 0.01$
& $0.25 \pm 0.04$ \\
\cmidrule(lr){1-9}
A2Summ
& $27.99 \pm 6.63$
& $0.00 \pm 0.00$
& $0.57 \pm 0.03$
& $1.01 \pm 0.04$
& $0.22 \pm 0.15$
& $0.13 \pm 0.02$
& $0.18 \pm 0.01$
& $0.24 \pm 0.04$ \\
TeaserGen \citep{xu2024teasergen}
& $4.30 \pm 0.23$
& $17.64 \pm 3.56$
& $0.56 \pm 0.04$
& $1.39 \pm 0.03$
& $0.06 \pm 0.14$
& $0.16 \pm 0.01$
& $0.18 \pm 0.01$
& $0.24 \pm 0.04$ \\
REGen \citep{xu2025regen}
& $3.37 \pm 0.86$
& $39.97 \pm 16.51$
& $0.53 \pm 0.04$
& $0.41 \pm 0.23$
& $0.26 \pm 0.26$
& $0.14 \pm 0.03$
& $\mathbf{0.19 \pm 0.01}$
& $0.26 \pm 0.04$ \\
\cmidrule(lr){1-9}
Clips2Story-NF
& $31.18 \pm 3.72$
& $0.34 \pm 0.24$
& $0.58 \pm 0.03$
& $1.03 \pm 0.03$
& $0.58 \pm 0.10$
& $0.16 \pm 0.02$
& $\mathbf{0.19 \pm 0.01}$
& $0.26 \pm 0.04$ \\
Clips2Story-ND
& $12.07 \pm 1.21$
& $10.79 \pm 1.26$
& $0.56 \pm 0.02$
& $1.22 \pm 0.03$
& $0.19 \pm 0.09$
& $\mathbf{0.18 \pm 0.02}$
& $\mathbf{0.19 \pm 0.01}$
& $\mathbf{0.27 \pm 0.04}$ \\
\bottomrule
\end{tabular}%
}
\end{table*}

\begin{table*}[!htbp]
\centering
\small
\setlength{\tabcolsep}{3pt}
\caption{Performance of A2Summ, TeaserGen, REGen, Clips2Story-NF, and Clips2Story-ND across five video genres under the clip-level setting, compared with the input video clips. Values are reported with 95\% confidence intervals.}
\label{tab:cross_genre_clip_level}
\vspace{1ex}
\resizebox{\textwidth}{!}{%
\begin{tabular}{lcccccccc}
\toprule
Model
& MCD (sec)
& REP (\%)
& CLIPScore
& VTGHLS
& OPC
& KR-Narration $\uparrow$
& KR-Frame $\uparrow$
& KR-Caption $\uparrow$ \\
\midrule

\multicolumn{9}{c}{\textbf{Documentary}} \\
\midrule
Input video clips
& $4.13 \pm 0.39$
& $0.00 \pm 0.00$
& $0.49 \pm 0.01$
& $1.02 \pm 0.03$
& $1.00 \pm 0.00$
& $0.17 \pm 0.02$
& $0.17 \pm 0.01$
& $0.23 \pm 0.02$ \\
\cmidrule(lr){1-9}
A2Summ
& $3.87 \pm 0.41$
& $0.00 \pm 0.00$
& $0.50 \pm 0.01$
& $1.07 \pm 0.02$
& $0.01 \pm 0.01$
& $0.21 \pm 0.02$
& $0.17 \pm 0.01$
& $0.23 \pm 0.02$ \\
TeaserGen \citep{xu2024teasergen}
& $3.48 \pm 0.11$
& $15.18 \pm 3.37$
& $0.65 \pm 0.01$
& $1.46 \pm 0.02$
& $0.11 \pm 0.06$
& $0.21 \pm 0.02$
& $0.19 \pm 0.01$
& $0.24 \pm 0.02$ \\
REGen \citep{xu2025regen}
& $3.69 \pm 0.23$
& $9.80 \pm 10.58$
& $0.62 \pm 0.02$
& $0.94 \pm 0.19$
& $0.02 \pm 0.07$
& $0.23 \pm 0.02$
& $\mathbf{0.20 \pm 0.01}$
& $0.23 \pm 0.03$ \\
\cmidrule(lr){1-9}
Clips2Story-NF
& $5.38 \pm 0.51$
& $0.38 \pm 0.31$
& $0.55 \pm 0.02$
& $1.13 \pm 0.03$
& $0.29 \pm 0.08$
& $0.31 \pm 0.02$
& $0.19 \pm 0.01$
& $0.25 \pm 0.03$ \\
Clips2Story-ND
& $4.28 \pm 0.24$
& $16.13 \pm 1.78$
& $0.57 \pm 0.01$
& $1.29 \pm 0.02$
& $0.13 \pm 0.04$
& $\mathbf{0.33 \pm 0.02}$
& $\mathbf{0.20 \pm 0.01}$
& $\mathbf{0.27 \pm 0.03}$ \\
\midrule

\multicolumn{9}{c}{\textbf{Film}} \\
\midrule
Input video clips
& $9.66 \pm 1.83$
& $0.00 \pm 0.00$
& $0.47 \pm 0.00$
& $0.90 \pm 0.02$
& $1.00 \pm 0.00$
& $0.11 \pm 0.01$
& $0.16 \pm 0.01$
& $0.20 \pm 0.02$ \\
\cmidrule(lr){1-9}
A2Summ
& $9.29 \pm 1.75$
& $0.00 \pm 0.00$
& $0.49 \pm 0.01$
& $0.95 \pm 0.01$
& $0.01 \pm 0.01$
& $0.12 \pm 0.01$
& $0.17 \pm 0.01$
& $0.20 \pm 0.02$ \\
TeaserGen \citep{xu2024teasergen}
& $3.95 \pm 0.20$
& $7.65 \pm 2.24$
& $0.59 \pm 0.02$
& $1.42 \pm 0.01$
& $0.04 \pm 0.06$
& $0.16 \pm 0.02$
& $0.17 \pm 0.01$
& $0.19 \pm 0.02$ \\
REGen \citep{xu2025regen}
& $4.48 \pm 0.70$
& $19.87 \pm 12.77$
& $0.53 \pm 0.04$
& $0.85 \pm 0.14$
& $0.08 \pm 0.16$
& $0.15 \pm 0.03$
& $\mathbf{0.18 \pm 0.01}$
& $0.18 \pm 0.03$ \\
\cmidrule(lr){1-9}
Clips2Story-NF
& $10.76 \pm 1.54$
& $0.22 \pm 0.17$
& $0.50 \pm 0.01$
& $0.99 \pm 0.02$
& $0.27 \pm 0.06$
& $0.19 \pm 0.02$
& $\mathbf{0.18 \pm 0.01}$
& $0.21 \pm 0.02$ \\
Clips2Story-ND
& $6.17 \pm 0.51$
& $8.99 \pm 0.93$
& $0.52 \pm 0.01$
& $1.22 \pm 0.01$
& $0.17 \pm 0.03$
& $\mathbf{0.22 \pm 0.02}$
& $\mathbf{0.18 \pm 0.01}$
& $\mathbf{0.22 \pm 0.02}$ \\
\midrule

\multicolumn{9}{c}{\textbf{Lecture}} \\
\midrule
Input video clips
& $165.22 \pm 129.49$
& $0.01 \pm 0.01$
& $0.50 \pm 0.02$
& $1.02 \pm 0.02$
& $1.00 \pm 0.00$
& $0.21 \pm 0.02$
& $0.19 \pm 0.01$
& $0.15 \pm 0.02$ \\
\cmidrule(lr){1-9}
A2Summ
& $31.32 \pm 14.68$
& $0.00 \pm 0.00$
& $0.57 \pm 0.04$
& $1.06 \pm 0.05$
& $0.17 \pm 0.14$
& $0.23 \pm 0.02$
& $0.19 \pm 0.01$
& $0.17 \pm 0.03$ \\
TeaserGen \citep{xu2024teasergen}
& $4.72 \pm 0.45$
& $31.86 \pm 5.99$
& $0.61 \pm 0.04$
& $1.36 \pm 0.07$
& $0.13 \pm 0.21$
& $0.22 \pm 0.02$
& $\mathbf{0.21 \pm 0.01}$
& $0.26 \pm 0.05$ \\
REGen \citep{xu2025regen}
& $7.40 \pm 1.73$
& $19.38 \pm 14.61$
& $0.58 \pm 0.05$
& $0.42 \pm 0.22$
& $0.32 \pm 0.19$
& $0.21 \pm 0.04$
& $\mathbf{0.21 \pm 0.01}$
& $\mathbf{0.27 \pm 0.06}$ \\
\cmidrule(lr){1-9}
Clips2Story-NF
& $23.20 \pm 5.87$
& $0.38 \pm 0.42$
& $0.54 \pm 0.06$
& $0.99 \pm 0.11$
& $0.19 \pm 0.25$
& $0.27 \pm 0.04$
& $0.20 \pm 0.01$
& $0.19 \pm 0.04$ \\
Clips2Story-ND
& $10.68 \pm 2.04$
& $20.09 \pm 8.08$
& $0.56 \pm 0.03$
& $1.20 \pm 0.04$
& $0.17 \pm 0.16$
& $\mathbf{0.32 \pm 0.03}$
& $\mathbf{0.21 \pm 0.01}$
& $0.22 \pm 0.03$ \\
\midrule

\multicolumn{9}{c}{\textbf{News}} \\
\midrule
Input video clips
& $8.44 \pm 1.96$
& $0.05 \pm 0.01$
& $0.53 \pm 0.03$
& $1.05 \pm 0.03$
& $1.00 \pm 0.00$
& $0.13 \pm 0.02$
& $0.15 \pm 0.01$
& $0.17 \pm 0.02$ \\
\cmidrule(lr){1-9}
A2Summ
& $8.92 \pm 1.84$
& $0.00 \pm 0.00$
& $0.58 \pm 0.04$
& $1.13 \pm 0.03$
& $0.02 \pm 0.06$
& $0.15 \pm 0.02$
& $0.16 \pm 0.01$
& $0.18 \pm 0.02$ \\
TeaserGen \citep{xu2024teasergen}
& $4.12 \pm 0.25$
& $16.14 \pm 3.92$
& $0.59 \pm 0.03$
& $1.41 \pm 0.03$
& $0.03 \pm 0.11$
& $0.14 \pm 0.02$
& $0.16 \pm 0.01$
& $0.20 \pm 0.02$ \\
REGen \citep{xu2025regen}
& $4.87 \pm 0.75$
& $9.44 \pm 7.42$
& $0.57 \pm 0.04$
& $1.03 \pm 0.17$
& $-0.03 \pm 0.11$
& $0.16 \pm 0.03$
& $0.17 \pm 0.01$
& $0.19 \pm 0.03$ \\
\cmidrule(lr){1-9}
Clips2Story-NF
& $10.41 \pm 2.79$
& $1.89 \pm 1.18$
& $0.58 \pm 0.03$
& $1.12 \pm 0.04$
& $0.37 \pm 0.06$
& $0.25 \pm 0.02$
& $\mathbf{0.20 \pm 0.01}$
& $0.22 \pm 0.03$ \\
Clips2Story-ND
& $6.19 \pm 1.07$
& $22.11 \pm 3.60$
& $0.55 \pm 0.03$
& $1.25 \pm 0.02$
& $0.13 \pm 0.03$
& $\mathbf{0.27 \pm 0.02}$
& $\mathbf{0.20 \pm 0.01}$
& $\mathbf{0.23 \pm 0.03}$ \\
\midrule

\multicolumn{9}{c}{\textbf{Vlog}} \\
\midrule
Input video clips
& $74.36 \pm 62.64$
& $0.02 \pm 0.01$
& $0.53 \pm 0.02$
& $0.99 \pm 0.03$
& $1.00 \pm 0.00$
& $0.13 \pm 0.02$
& $0.18 \pm 0.01$
& $0.25 \pm 0.04$ \\
\cmidrule(lr){1-9}
A2Summ
& $30.40 \pm 6.97$
& $0.00 \pm 0.00$
& $0.58 \pm 0.02$
& $1.03 \pm 0.04$
& $0.05 \pm 0.07$
& $0.13 \pm 0.02$
& $0.18 \pm 0.01$
& $0.25 \pm 0.04$ \\
TeaserGen \citep{xu2024teasergen}
& $4.30 \pm 0.15$
& $13.48 \pm 4.64$
& $0.57 \pm 0.04$
& $1.38 \pm 0.03$
& $0.05 \pm 0.14$
& $0.17 \pm 0.01$
& $0.19 \pm 0.01$
& $0.26 \pm 0.04$ \\
REGen \citep{xu2025regen}
& $4.08 \pm 0.83$
& $42.58 \pm 19.27$
& $0.58 \pm 0.05$
& $0.66 \pm 0.25$
& $-0.04 \pm 0.22$
& $0.16 \pm 0.04$
& $0.19 \pm 0.02$
& $0.26 \pm 0.06$ \\
\cmidrule(lr){1-9}
Clips2Story-NF
& $26.36 \pm 3.71$
& $0.42 \pm 0.42$
& $0.58 \pm 0.03$
& $1.02 \pm 0.03$
& $0.37 \pm 0.11$
& $0.16 \pm 0.02$
& $\mathbf{0.20 \pm 0.01}$
& $0.27 \pm 0.04$ \\
Clips2Story-ND
& $10.89 \pm 1.15$
& $11.68 \pm 1.57$
& $0.56 \pm 0.02$
& $1.23 \pm 0.03$
& $0.12 \pm 0.09$
& $\mathbf{0.18 \pm 0.01}$
& $0.19 \pm 0.01$
& $\mathbf{0.28 \pm 0.04}$ \\
\bottomrule
\end{tabular}%
}
\end{table*}



\end{document}